\documentclass[10pt,journal,compsoc]{IEEEtran}

\ifCLASSOPTIONcompsoc
   \usepackage{cite}
\else
  \usepackage{cite}
\fi

\ifCLASSINFOpdf
   \usepackage[pdftex]{graphicx}
   \usepackage{epstopdf}
\else
  
  \usepackage[dvips]{graphicx}

  \DeclareGraphicsExtensions{.eps}

\fi

\usepackage{amsmath}
\usepackage{amsthm}
\usepackage{multirow}
\usepackage{color}
\usepackage{amssymb}
\usepackage{algorithm}
\usepackage{algorithmic}
\usepackage{array}
\usepackage{bm}
\usepackage{hyperref}       % hyperlinks
\usepackage{url}            % simple URL typesetting
\usepackage{booktabs}       % professional-quality tables
\usepackage{amsfonts}       % blackboard math symbols
\usepackage{nicefrac}       % compact symbols for 1/2, etc.
\usepackage{microtype}      % microtypography
\usepackage[dvipsnames, svgnames, x11names]{xcolor}      % colors
\usepackage{algorithm} %format of the algorithm 
\usepackage{algorithmic} %format of the algorithm 
\usepackage{threeparttable}
\usepackage[switch]{lineno}

\usepackage{subfigure}

\newtheorem{Assumption}{Property}

\begin{document}
%\linenumbers
%

\title{Background-aware Classification Activation Map for Weakly Supervised Object Localization}

\author{Lei~Zhu, Qi~She, Qian~Chen, Xiangxi~Meng, Mufeng~Geng, Lujia~Jin, Zhe~Jiang, Bin~Qiu, Yunfei~You, Yibao~Zhang, Qiushi~Ren, Yanye~Lu*% <-this % stops a space<-this % stops a space
\thanks{This work was supported by the Beijing Natural Science Foundation under Grant Z210008, in part by Shenzhen Science and Technology Program under Grant 1210318663, and in part by Shenzhen Nanshan Innovation and Business Development Grant. }
        \thanks{Lei~Zhu, Qian~Chen, Mufeng~Geng, Lujia~Jin, Zhe~Jiang, Bin~Qiu, Yunfei~You, Qiushi~Ren are with the Institute of Medical Technology, Peking University Health Science Center, Peking University, Beijing 100191, China, also with the Department of Biomedical Engineering, Peking University, Beijing 100871, China, also with the Institute of Biomedical Engineering, Shenzhen Bay Laboratory, Shenzhen 5181071, China, and also with the Institute of Biomedical Engineering, Peking University Shenzhen Graduate School, Shenzhen 518055.}
        \thanks{Yanye Lu is with the Institute of Medical Technology, Peking University Health Science Center, Peking University, Beijing 100191, China, and also with the Institute of Biomedical Engineering, Peking University Shenzhen Graduate School, Shenzhen 518055, China, email: yanye.lu@pku.edu.cn}
        \thanks{Qi~She is with the ByteDance AI Lab, ByteDance, Beijing 100086, China.}
        \thanks{Xiangxi~Meng, Yibao~Zhang is with the Key Laboratory of Carcinogenesis and Translational Research (Ministry of Education), Peking University Cancer Hospital \& Institute, Beijing 100142, China}% <-this % 
}

\markboth{Journal of \LaTeX\ Class Files,~Vol.~14, No.~8, August~2015}%
{Shell \MakeLowercase{\textit{et al.}}: Bare Demo of IEEEtran.cls for Computer Society Journals}

\IEEEtitleabstractindextext{%
\begin{abstract}
Weakly supervised object localization (WSOL) relaxes the requirement of dense annotations for object localization by using image-level classification masks to supervise its learning process. However, current WSOL methods suffer from excessive activation of background locations and need post-processing to obtain the localization mask. This paper attributes these issues to the unawareness of background cues, and propose the background-aware classification activation map (B-CAM) to simultaneously learn localization scores of both object and background with only image-level labels. In our B-CAM, two image-level features, aggregated by pixel-level features of potential background and object locations, are used to purify the object feature from the object-related background and to represent the feature of the pure-background sample, respectively. Then based on these two features, both the object classifier and the background classifier are learned to determine the binary object localization mask. Our B-CAM can be trained in end-to-end manner based on a proposed stagger classification loss, which not only improves the objects localization but also suppresses the background activation. Experiments show that our B-CAM outperforms one-stage WSOL methods on the CUB-200, OpenImages and VOC2012 datasets.
\end{abstract}
}

\maketitle

\IEEEdisplaynontitleabstractindextext
\IEEEpeerreviewmaketitle

%introduction
\IEEEraisesectionheading{\section{Introduction}\label{sec:introduction}}
\IEEEPARstart{W}{eakly} supervised learning (WSL), using minimal supervision or coarse annotations for model learning, has attracted extensive attention in recent years and has been widely used in computer vision tasks~\cite{WSIS, WSVSVL, SEAM, CAM, WSR}. Among them, weakly supervised object localization (WSOL) has immensely profited from WSL, where the requirement of location annotations such as pixel-level masks or bounding boxes can be replaced by easily obtained image-level classification labels. It usually adopts the flow of classification activation map (CAM)~\cite{CAM} that utilizes the structure of image classification to generate the localization score via appending a global average pooling (GAP) operation and a fully connected layer after the feature extractor, \textit{i.e.}, the convolutional network. 

Unfortunately, when used for the WSOL tasks, CAM usually activates the most discriminative object part rather than the whole object and requires post-processing to generate the localization mask. A series of WSOL methods have been developed to overcome these issues, which can be divided into multi-stage~\cite{UPSP, PSOL, GC, STL} and one-stage~\cite{HAS, ACOL, ADL, CUTMIX, SPG, CSOA, CAAM, SEM} methods. The former involve additional training stages as pre- or post-processing to enhance the quality of the localization map or generate class-agnostic localization results, which seriously increases the complexity of both the training and the test processes. While, the latter usually adopt different data-augmentation strategies~\cite{HAS,ADL,CUTMIX, ACOL} to erase discriminative object parts or use the coarse pixel-level mask as additional pixel-level supervision~\cite{SPG, CSOA, CAAM, SEM} to enhance the activation of undiscriminating parts of the objects. Though one-stage methods are more efficient and easy to train, they require post-thresholding to generate localization masks.

\begin{figure}
\centering
\includegraphics[width=0.49\textwidth]{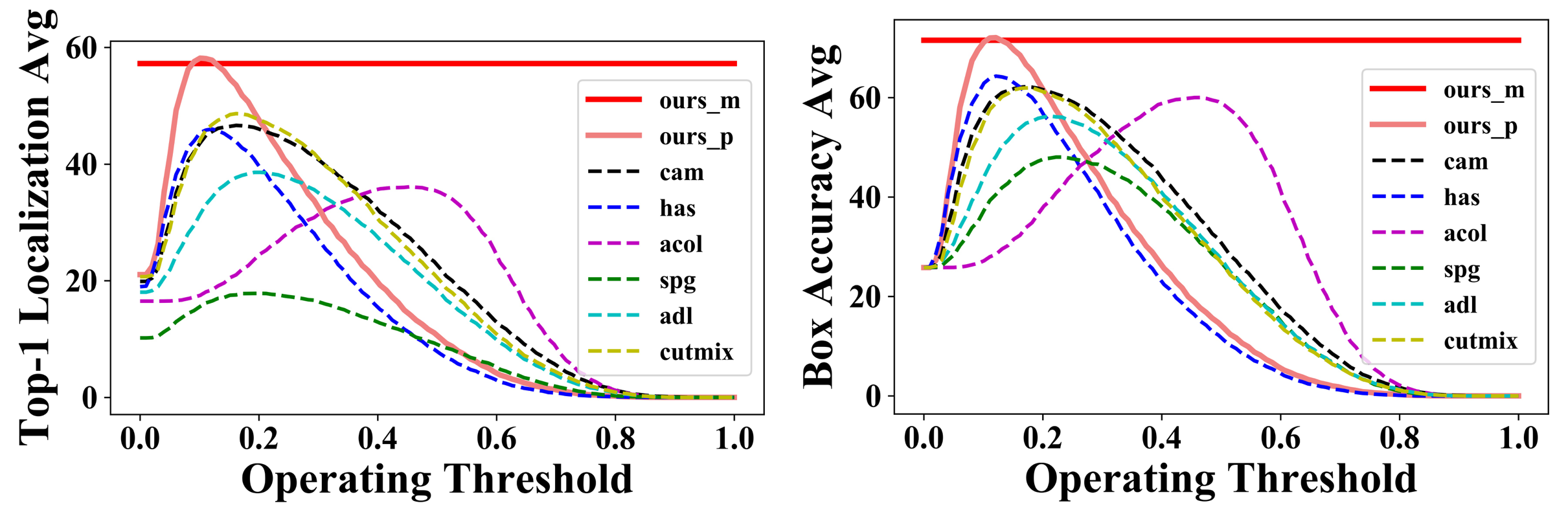}
\caption{The background thresholding step significantly affects the performance of one-stage WSOL methods. Our work attempts to solve this problem by directly generating pixel-level background score and object score simultaneously based on a background classifier.}
\label{fig:intro_perform}
\end{figure}

In contrast to the existing works, we attributed the above two problems of the CAM to its unawareness of the background. Specifically, in the object localization task, an input image must contain at least one object, making the pure-background class remain ``unseen'' for the image-label-supervised WSOL task. As the result, CAM only has the ability to discern different object classes based on image-level features, but cannot simultaneously identify whether the location belongs to object parts or background stuff. Thus, additional stages or post-thresholding is required for the CAM to generate background localization scores to fill this gap between localization and classification. As shown in Fig.~\ref{fig:intro_perform}, such post-processing influences the performance of one-stage WSOL methods to a great extent.

\begin{figure}
\centering
\includegraphics[width=0.49\textwidth]{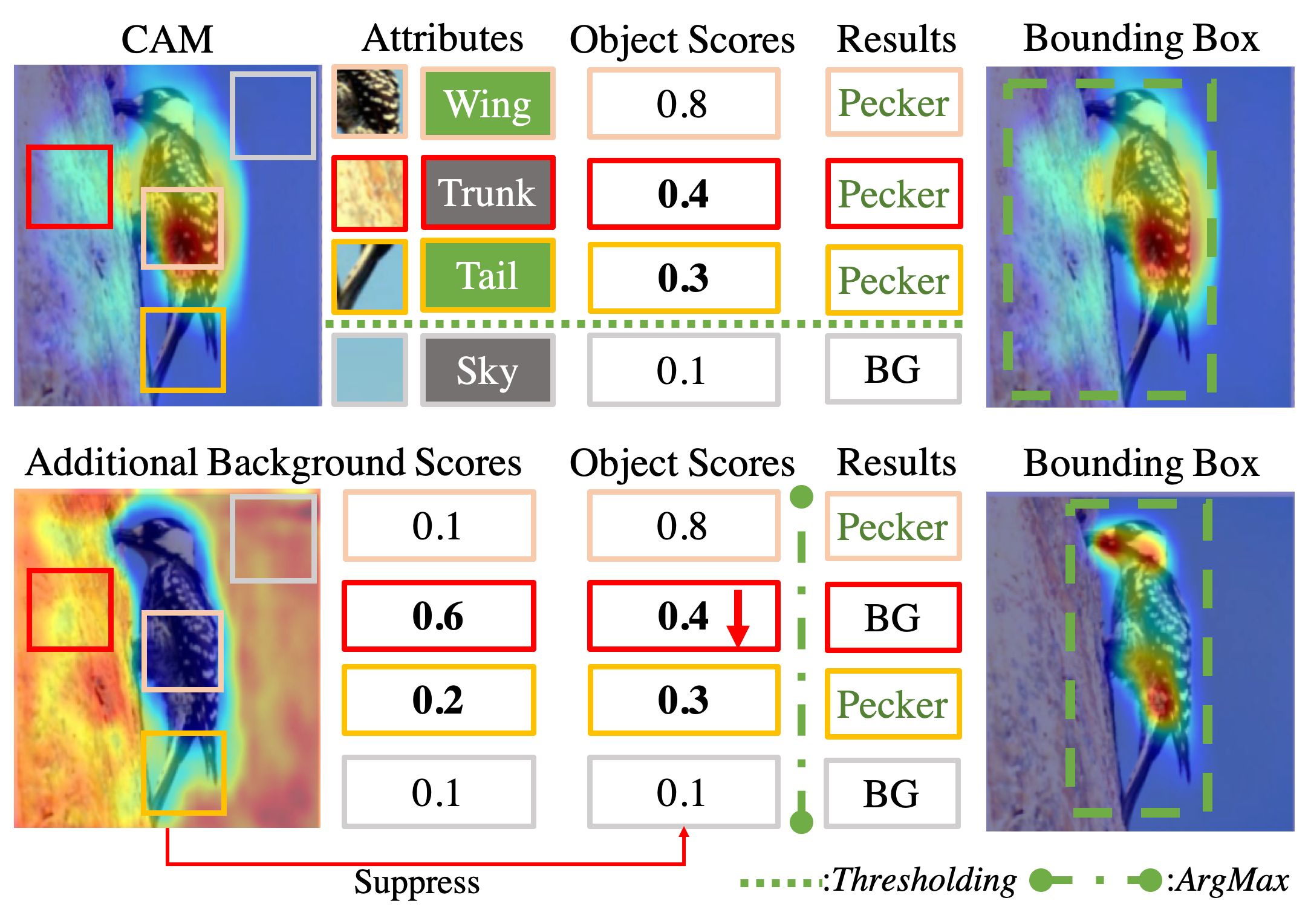}
\caption{Activation of object-related background limits the upper bound of WSOL. Our method generates pixel-level background scores to replace the image-level threshold and suppress the background activations. }
\label{fig:intro}
\end{figure}

Moreover, the unawareness of background also prevents the CAM from suppressing the excessive activation of the background locations~\cite{Survey}, especially the object-related background that is also discriminative to discern the object. For example, in the first row of Fig.~\ref{fig:intro}, the background ``trunk'' is also discriminative for discerning ``woodpecker'', resulting in a higher activation score in locations of ``trunk'' relative to ''the tail of the bird''. Thus, even using the optimal threshold, the tail of the bird will still be assigned to the ``background trunk'' rather than the ``foreground woodpecker''. If the background cues can be perceived as in the second row of Fig.~\ref{fig:intro}, the excessive background activation can be suppressed with the help of this background awareness, and the upper bound performance of WSOL methods can be improved.

In this paper, we focus on adding background awareness for one-stage WSOL by proposing an end-to-end learning structure called the background-aware classification activation map (B-CAM). Our B-CAM attempts to create the ``unseen pure-background samples'' by aggregating the potential locations of the background stuff and the foreground object respectively, rather than simply aggregating features with the GAP layer. Then, based on the aggregation features, an additional background classifier can be simultaneously learned with the object classifier by considering background prediction as a multi-label classification task. Moreover, the background aggregation features are also utilized to represent pure-background samples for the object estimator, suppressing background activation on object prediction with the help of a proposed stagger classification (SC) loss.

In a nutshell, our contributions are fourfold:
\begin{itemize}
\item {To our knowledge, our paper is the first one-stage WSOL work that simultaneously learns both object and background classifiers with image-level labels.}
\item {A novel structure called B-CAM is presented, generating pixel-level background scores and suppressing the background activation on the localization map.}
\item {An SC loss is elaborated to efficiently train our B-CAM with image-level labels in an end-to-end manner.}
\item{Experiments indicate that our method outperforms other one-stage methods with comparable complexity.}
\end{itemize}

%%%%%%

\section{Related Work}

\subsection{One-stage Weakly Supervised Object Localization}
One-stage WSOL methods follow the pipeline of CAM~\cite{CAM}, which directly adopts the classification structure to generate localization score by projecting the classification head (object estimator) back to the pixel-level feature map. However, due to the absence of localization supervision, CAM cannot effectively catch the indiscriminating parts of objects. To solve this problem, some one-stage WSOL methods focused on applying augmentation strategies to input images or feature maps to erase the discriminative object parts. Yun \textit{et al.}~\cite{CUTMIX} proposed a CutMix augmentation strategy, which replaces a patch of an image with another image to force the model to capture the indiscriminative parts. Singh \textit{et al.}~\cite{HAS} randomly hid the patches of images in the training process to discover different object parts. Zhang \textit{et al.}~\cite{ACOL} then simplified this augmentation by proposing an end-to-end network that contains two adversarial classifiers to complementarily capture object parts. Junsuk \textit{et al.}~\cite{ADL, ADL_conf} further adopted the attention mechanism to drop the discriminative parts of the feature map. Though these methods can capture more parts of the objects, they inevitably increase the activation of background, especially the object-related background that also contributes to determining the class of objects.
%%%

Apart from adopting the augmentation strategies, some one-stage WSOL methods also attempt to use coarse pixel-level supervision to train the object estimator. Zhang \textit{et al.}~\cite{SPG} proposed the self-produced guidance (SPG) approach, which generates an auxiliary pixel-level mask based on the attention map of different extractor stages to perceive background cues. Kou \textit{et al.}~\cite{CSOA} further generalized SPG by adding an additional object estimator to adaptively produce the auxiliary pixel-level mask, which is then utilized to design a metric learning loss to better supervise the training process. Ki \textit{et al.}~\cite{ICLCA} focused on enlarging the distance between features of object locations and background locations in the latent space with the help of the coarse mask generated by non-local attention. Babar \textit{et al.}~\cite{CAAM} attempted to enhance the localization map by aligning the localization scores of two complementary images, where these two scores supervise each other in pixel-level. However, the additional pixel-level supervision increases the complexity of  the training process, making them quite hard to train.

In contrast to the one-stage WSOL methods above, our B-CAM only uses image-level labels in the training process to perceive background cues rather than using additional pixel-level supervision. Moreover, our B-CAM also avoids the post-thresholding step that is required by other one-stage WSOL methods without using any additional stages.

\subsection{Multi-stage Weakly Supervised Object Localization}

Multi-stage WSOL methods add additional pre- or post-stages upon the classification structure to pursue better localization scores. Some multi-stage WSOL methods were elaborated to enhance the localization map of the one-stage WSOL by proposing novel post-processing. Zhang \textit{et al.}~\cite{SEM} added an additional learning-free-post-stage upon CAM to generate the self-enhanced map (SEM), which explores the correlation between each location and the seeds (locations with high localization scores). Pan \textit{et al.}~\cite{UPSP} further extended this approach by considering both first- and second-order self-correlation when aggregating the enhanced localization map. Though these methods can enhance the quality of localization maps, they still require post-thresholding to generate the background scores. 

Some other multi-stage WSOL methods focus on generating class-agnostic localization masks by the additional stages. The most typical work is the pseudo supervised object localization (PSOL) proposed by Zhang \textit{et al.}~\cite{PSOL}. In addition to the classification stage, PSOL also adds two additional stages to generate localization results. In the first additional stage (localizer), the one-stage WSOL method is trained to generate the coarse class-agnostic bounding boxes. Then in the second stage (regressor), those coarse boxes are used as the ground-truth to fully-supervised train bounding boxes regression to generate the region of interest-objects (ROI), i.e. the localization bounding-box. Guo \textit{et al.}~\cite{STL} improved PSOL by using a class-tolerance classification model for the localizer to enhance the quality of the coarse bounding boxes. However, these two methods can not generate pixel-level localization masks as one-stage WSOL methods. As a replacement, another three-stage WSOL method proposed by Lu \textit{et al.}~\cite{GC} adopts a generator, implemented by learning- or model-driven approaches, to generate class-agnostic binary masks based on the ROI with different geometry shapes (for example rectangle or ellipse). In addition, a detector and classifier are also trained to generate the ROI and class of objects, respectively. Though these multi-stage WSOL methods can directly generate localization results without post-thresholding, their additional stages increase both time and space complexity of the training process. %More recently, Meng~\textit{et al.}~\cite{FAM} improve the multi-stage WSOL methods by jointly optimizing class-agnostic localization and classification to pursue better localization results. Wu~\textit{et al.}~\cite{BAS} extend it by suppressing the background activation on the class-agnostic localization map during the jointly optimizing process. However, these two methods still need post-thresholding to generate localization mask.

Compared with these multi-stage methods, our B-CAM learns the background  classifier and object classifier simultaneously, rather than adopting additional training stages for class-agnostic localization. Moreover, both the object and background scores generated by our B-CAM are not class-agnostic, which enhances the flexibility when engaging in multi-object localization.

\subsection{Background Effect in Weakly Supervised Learning}

There are also some weakly supervised-learning methods in other scopes designed to capture background cues. Oh \textit{et al.}~\cite{BAP} proposed a background-aware pooling strategy for the weakly supervised semantic segmentation (WSSS) with bounding-boxes annotations, which uses the region out of the ground-truth bounding boxes to catch the inner-boxes background locations. Lee \textit{et al.}~\cite{EPS} utilized the additional saliency map as pixel-level supervision to perceive background cues and reserve rich boundaries for WSSS. Fan \textit{et al.}~\cite{ICD} generated background scores for each class by learning intra-class boundaries, which requires additional superpixel and coarse pixel-level mask during network training. Lee \textit{et al.}~\cite{WSAL} proposed two background-aware losses that suppress the localization score of the background frame in the weakly supervised action localization (WSAL).

Unlike these methods, our B-CAM is designed for WSOL tasks that is harder to locate background cues than WSAL. Moreover, our B-CAM can perceive the background cues through only image-level labels rather than using the additional pixel-level supervision or off-the-shelf process, for example, the object proposal~\cite{RP}, saliency detection~\cite{SALIENCY}, superpixel segmentation~\cite{SLIC}, or conditional random fields~\cite{CRF}.

\section{Methodology}

\begin{figure*}
\centering
\includegraphics[width=0.99\textwidth]{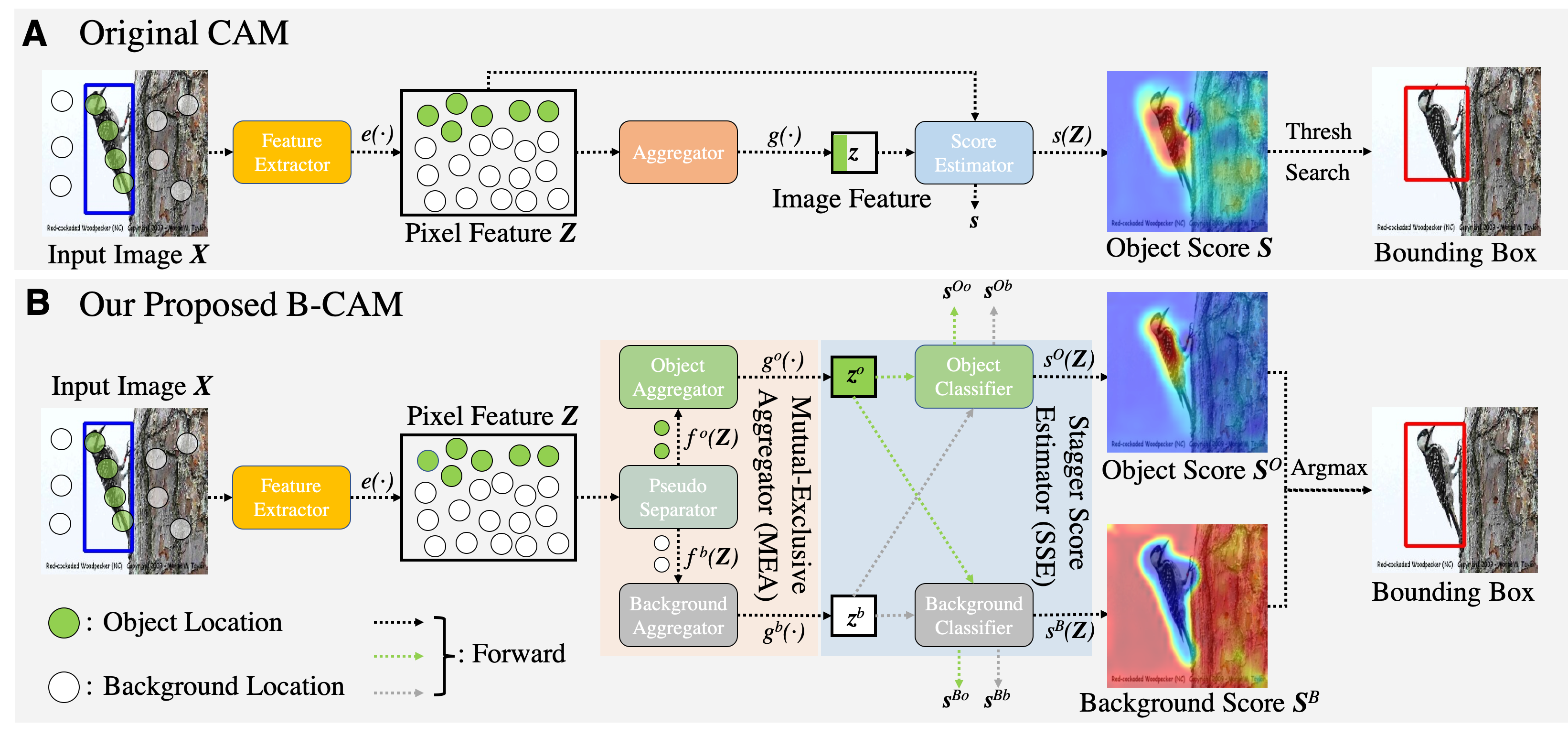}
\caption{The comparison of CAM and our B-CAM. \textbf{\sf{A}}: the illustration of CAM. \textbf{\sf{B}}: the illustration of our B-CAM that replaces the feature aggregator and score estimator by the proposed MEA and SSE. }
\label{fig:overview}
\end{figure*}

In this section, we first provide the formulation of the WSOL task and review the CAM pipeline adopted by existing WSOL methods to analyze why it cannot perceive background cues. Then, we illustrate the strategy of our proposed B-CAM, which adds background awareness with only image-level supervision. Next, we introduce the network structure of our B-CAM in detail, including the proposed mutual-exclusive aggregator (MEA) and stagger score estimator (SSE). Finally, we propose the stagger classification (SC) loss that supervises the training process of our B-CAM. 

%Note that, we use \textbf{bold} uppercase characters to denote the matrix-valued random variables (the parameter matrices) and \textit{\textbf{italic bold}} uppercase characters to denote other matrices (such as feature maps). Vectors are denoted with lowercase. 

\subsection{Revisiting Weakly Supervised Object Localization}

Given an input image represented by a matrix $\bm{X} \in \mathbb{R}^{3 \times N}$, the object localization task uses a scoring function $p(\bm{X})$ to approximate the binary mask $\bm{Y} \in \mathbb{R}^{K \times N}$, whose element $\bm{Y}_{k, i}$ identifies whether or not pixel $i$ belongs to the object of a specific class $k$. Note that $K$ and $N$ are the numbers of classes of interest and pixels, respectively. In contrast to the fully supervised object localization that uses $\bm{Y}$ to supervise the learning of $p(\bm{X})$ at pixel-level, WSOL refers to the special condition that only the image-level mask $\bm{y} = \big(\max(\bm{Y}_{0, :}), \max(\bm{Y}_{1, :}), ..., \max(\bm{Y}_{K-1, :})\big) \in \mathbb{R}^{K \times 1}$ is available for the whole training process.

Existing one-stage WSOL methods follow the pipeline of CAM~\cite{CAM} that splits the scoring function $p(\bm{X})$ into three subparts to learn it with $\bm{y}$. As shown in Fig.~\ref{fig:overview}~\textbf{\sf{A}}, it mainly contains three parts: (1) feature extractor $e(\cdot)$ uses a backbone network to extract pixel-level features $\bm{Z} \in \mathbb{R}^{C \times N}$, where $C$ is the number of channels; (2) feature aggregator $g(\cdot)$ utilizes GAP~\cite{GAP} to generate an image-level feature $\bm{z} \in \mathbb{R}^{C \times 1}$; (3) score estimator $s(\cdot)$ adopts a fully connected layer to obtain the image-level classification score $\bm{s} = s(\bm{z}) \in \mathbb{R}^{K \times 1}$, which can be projected onto the pixel-level features to obtain the localization score $\bm{S} = \big( s(\bm{Z}_{:, 0}), s(\bm{Z}_{:, 1}), ..., s(\bm{Z}_{:, N-1}) \big) \in \mathbb{R}^{K \times N}$ during the test process. Based on this pipeline, the cross-entropy loss $\mathcal{L}_{ce}(\bm{y}, \bm{s})$ between  the classification score $\bm{s}$ and the ground-truth image-level mask $\bm{y}$ can supervise the training process.

\subsection{Cause of the Background Unawareness}

Though the CAM pipeline can learn the scoring function $p(\bm{X})$ with the image-level label $\bm{y}$, it suffers from excessive activation on background locations and needs post-thresholding to generate the background score, which limits its performance severely. We attribute these problems to the different targets between CAM and object localization, where CAM pays too much attention to the image-level object classification without concerning the characteristic of object localization, \textit{i.e.}, the background locations, which are also crucial and need to be discern for the localization task.

Firstly in CAM, the image-level feature used to estimate the classification score is contaminated by background features. As shown in Fig.~\ref{fig:overview}~\textbf{\sf{A}}, the GAP layer of the feature aggregator (noted by peach color), proposed for the image classification task, treats pixel-level features of the object and the background equally when summarizing the image representations. As a result, $\bm{z}$ is inevitably influenced by the distribution of the background, where some object-related background cues can also assist the classifier in discerning image classes, as in the case of the background ``trunk'' \textit{vs.} the object ``woodpecker''. Although this influence can improve the accuracy and interpretability for image classification, it causes undesirable background activation for WSOL that generates object localization scores by projecting the classifier (score estimator) back to the pixel-level features, where background locations are also contained.

Moreover, pure-background samples are not existed for the training of CAM in WSOL. Specifically, unlike the pixel-level binary mask $\bm{Y}$, the image-level classification mask $\bm{y}$ does not contain any samples that satisfy $\bm{y} = \bm{0}$, which are crucial for object localization to percept background locations. The absence of these samples not only diminishes the capacity in suppressing background activation for the score estimator , but also disables training a classifier to generate the background localization scores for WSOL task.

\begin{figure*}
\centering
\includegraphics[width=0.99\textwidth]{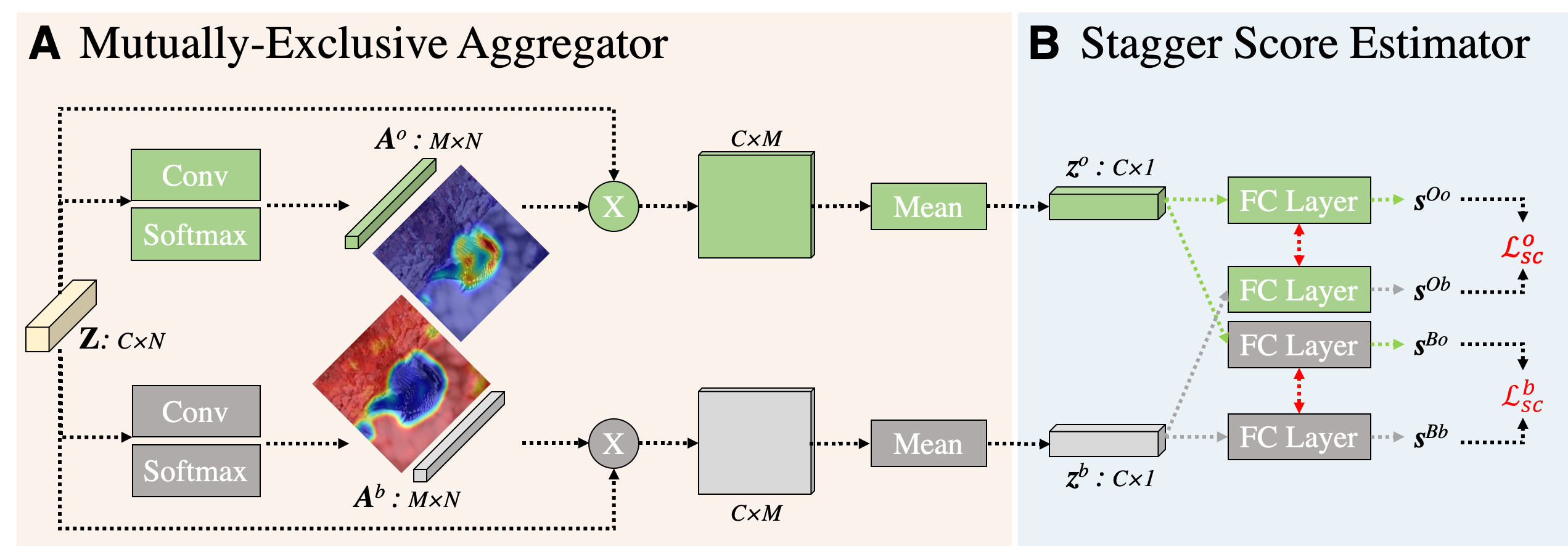}
\caption{The structure of the proposed modules of our B-CAM. \textbf{\sf{A}}: the structure of our MEA where ``$\times$'' means the multiplication operation and coarse priors are visualized by its row-wise sum. \textbf{\sf{B}}: the structure of our SSE implemented as two fully connected layer with stagger connection. Red line means weight sharing.}
\label{fig:structure}
\end{figure*}

\subsection{Background-aware Classification Activation Map}

%%%%%%%%%%

To solve the above factors, our B-CAM is proposed as shown in Fig.~\ref{fig:overview}~\textbf{\sf{B}}. In general, the key idea of our B-CAM is to replace the GAP-based aggregator with a self-trained aggregator called the mutual-exclusive aggregator (MEA). The proposed MEA generates two image-level features ($\bm{z}^{o}$ and $\bm{z}^{b}$) by aggregating features on the potential location of the object part and background part, respectively. To achieve this purpose, our MEA firstly uses two functions $f^{o}(\cdot)$ and $f^{b}(\cdot)$ to generate coarse object prior $\bm{A}^{o} \in \mathbb{R}^{M \times N}$ and background prior $\bm{A}^{b} \in \mathbb{R}^{M \times N}$ for each spatial position of the feature map $\bm{Z}$, where $M$ is a hyper-parameter. Then an aggregate function $g(\cdot)$ is adopted on $\bm{Z}$ subject to the two priors, \textit{i.e.} $\bm{A}^{o}$ and $\bm{A}^{b}$, to generate image-level features $\bm{z}^{o}$ and $\bm{z}^{b} \in \mathbb{R}^{C \times 1}$ for the object and the background, respectively. Profited by these two priors, the image-level object feature $\bm{z}^{o}$ is less contaminated by the features of background location than the image-level feature $\bm{z}$ of original CAM. Meanwhile, the additional image-level background feature $\bm{z}^{b}$ simulates the feature aggregated from ``the pure-background image''. This sample helps train an additional background classifier and suppress the background activation for the object classifier.

Our B-CAM also adopts a stagger score estimator (SSE), which adds an additional background classifier upon the score estimator (object classifier) of CAM to generate the background score. By feeding the two image-level features $\bm{z}^{o}$ and $\bm{z}^{b}$ into our SSE, four classification scores $\bm{s}^{Oo}, \bm{s}^{Ob}, \bm{s}^{Bo}$, and $\bm{s}^{Bb} \in \mathbb{R}^{K \times 1}$ can be obtained, which are all used to supervise the network learning with the help of the image-level mask $\bm{y}$. Specifically, the background classifier $s^{B}(\cdot)$ estimates two background classification scores $\bm{s}^{Bo}$ and $\bm{s}^{Bb}$ respectively for the image-level object feature $\bm{z}^{o}$ and image-level background feature $\bm{z}^{b}$. In a similar way, the object classifier $s^{O}(\cdot)$ generates two object classification scores $\bm{s}^{Oo}$ and $\bm{s}^{Ob}$. Profited by the specialty of these four classification scores, we finally elaborate a stagger classification (SC) loss $\mathcal{L}_{sc}(\cdot)$ to train our model with only the image-level mask $\bm{y}$. The SC loss not only guarantees the precision of both the object and the background classification scores but also suppresses the activation of negative samples. Note that the two classifiers of SSE can be projected back onto the pixel-level features to obtain the object localization score $\bm{S}^{O}$ and the background localization score $\bm{S}^{B}$ during the test process. Those two localization scores are then utilized to decide the final binary localization mask $\bm{Y}^{*}$. 

\subsection{Network design}
The structure of our B-CAM is composed of three parts: (1) the feature extractor; (2) the proposed MEA; (3) the proposed SSE. The feature extractor can be implemented by the classification backbones~\cite{RESNET, RESNET38} as the CAM. So only the proposed modules are elaborated in the following parts.

\subsubsection{Mutual-exclusive aggregator}
After obtaining the pixel-level feature map $\bm{Z}$ by the feature extractor $e(\cdot)$, our proposed B-CAM utilizes MEA to generate two image-level features for foreground objects and background stuff based on the coarse location priors $\bm{A}^{o}$ and $\bm{A}^{b} \in \mathbb{R}^{M \times N}$. Each column-vector of the priors can be viewed as a spatial attention map to capture the different spatial relations of objects/stuff by activating some potential locations that belongs to object (or background). As indicated in Fig.~\ref{fig:structure}~\textbf{\sf{A}}, a multi-head spatial attention structure is utilized to generate these two coarse location priors. In detail, the location prior $\bm{A}^{o}$ (or $\bm{A}^{b}$) are obtained by feeding the pixel-level feature $\bm{Z}$ into a convolution with column-wise softmax activation: 
\begin{equation} 
\left\{
\begin{aligned}
\bm{A}^{o}_{:, i} = f^{o}(\bm{Z})= \frac{\exp(\mathbf{W}_{1} * \bm{Z}_{:, i})}{\sum^{N}_{j}\exp(\mathbf{W}_{1} * \bm{Z}_{:, j})} \\
\bm{A}^{b}_{:, i} = f^{b}(\bm{Z}) = \frac{\exp(\mathbf{W}_{2} * \bm{Z}_{:, i})}{\sum^{N}_{j}\exp(\mathbf{W}_{2} * \bm{Z}_{:, j})}
\end{aligned}
\right. ,
\label{eq:A}
\end{equation}
\noindent where $\mathbf{W}_{1} \in \mathbb{R}^{M \times C}$ and $\mathbf{W}_{2} \in \mathbb{R}^{M \times C}$ are weight matrices to generate the object and background prior, respectively.

Then based on the coarse location priors $\bm{A}^{o}$ and $\bm{A}^{b}$, two aggregators are adopted to generate the image-level object feature $\bm{z}^{o}$ and background feature $\bm{z}^{b}$ with the help of the attention pooling strategy. As shown in Fig.~\ref{fig:structure}~\textbf{\sf{A}}, the corresponding column-vectors of $\bm{A}^{o}$ (or $\bm{A}^{b}$) are utilized as the attention map to pooling the pixel-level feature $\bm{Z}$ into $M$ different aggregation features. The mean strength of these $M$ image-level features is used as the final image-level features:
\begin{equation} 
\left\{
\begin{aligned}
\bm{z}^{o} = g(\bm{Z}, \bm{A}^{o}) = \frac{1}{M}\sum\nolimits_{m}^{M}\sum\nolimits_{i}^{N}{\bm{A}^{o}_{m, i}\bm{Z}_{:, i}} \\
\bm{z}^{b} = g(\bm{Z}, \bm{A}^{b}) = \frac{1}{M}\sum\nolimits_{m}^{M}\sum\nolimits_{i}^{N}{\bm{A}^{b}_{m, i}\bm{Z}_{:, i}}
\end{aligned}
\right. .
\label{eq:z}
\end{equation}

\subsubsection{Stagger score estimator}
Based on the two image-level features, SSE is adopted to generate the image-level classification scores, which are used to supervise the learning process of our B-CAM with the image-level mask $\bm{y}$. Specifically, as shown in Fig.~\ref{fig:structure}~\textbf{\sf{B}}, each image-level feature $\bm{z}^{o}$ (or $\bm{z}^{b}$) is fed into two independent score estimators implemented as fully connected layers to estimate whether the feature belongs to the foreground object sample or pure-background sample of each class:
\begin{equation} 
\left\{
\begin{aligned}
\bm{s}^{Oo} = s^O(\bm{z}^o) = \mathbf{W}_{O} * \bm{z}^{o} \\
\bm{s}^{Bo}  = s^B(\bm{z}^o) = \mathbf{W}_{B} * \bm{z}^{o} 
\end{aligned}
\right. ~,~
\left\{
\begin{aligned}
\bm{s}^{Ob} = s^O(\bm{z}^b) = \mathbf{W}_{O} * \bm{z}^{b} \\
\bm{s}^{Bb} = s^B(\bm{z}^b) = \mathbf{W}_{B} * \bm{z}^{b}
\end{aligned}
\right. ,
\label{eq:sse}
\end{equation}

\noindent where $\mathbf{W}_{o}$ and $\mathbf{W}_{B} \in \mathbb{R}^{K \times C}$ are two weight matrices of the fully connected layers that estimate the object or background scores for the classes of interest, respectively. $\bm{s}^{Oo}$ and $\bm{s}^{Bo}$ are the object classification score and background classification score of the  image-level object feature $\bm{z}^{o}$; $\bm{s}^{Ob}$ and $\bm{s}^{Bb}$ are the corresponding scores of the image-level background feature $\bm{z}^{b}$. These four localization scores are utilized by the SC loss to supervise the training process.

During the inference process, the SSE is projected back onto the pixel-level features $\bm{Z}$ to generate the pixel-level object and background localization scores:
\begin{equation} 
\bm{S}^{O}_{:, i} = \mathbf{W}_{O} * \bm{Z}_{:, i} ~~~,~~~ \bm{S}^{B}_{:, i} = \mathbf{W}_{B} * \bm{Z}_{:, i}, 
\label{eq:S}
\end{equation}
\noindent where $\bm{S}^{O}_{:, i}$ and $\bm{S}^{B}_{:, j}$ are the object and background localization scores of the $i^{th}$ pixel. Based on these two localization maps, the binary localization mask can be directly obtained:
\begin{equation}
\bm{Y}^{*}_{k, i} = \arg\max(\bm{S}^{B}_{k, i},  \bm{S}^{O}_{k, i}),
\label{eq:Y}
\end{equation}
\noindent where $\bm{Y}^{*}_{k, i}=1$ means pixel $i$ belongs to the object with class $k$. Thus, the localization mask $\bm{Y}^{*}$ of our B-CAM can be directly generated without using thresholding or additional stages as other WSOL methods.

\begin{figure}%[!htp]
\centering
\includegraphics[width=0.49\textwidth]{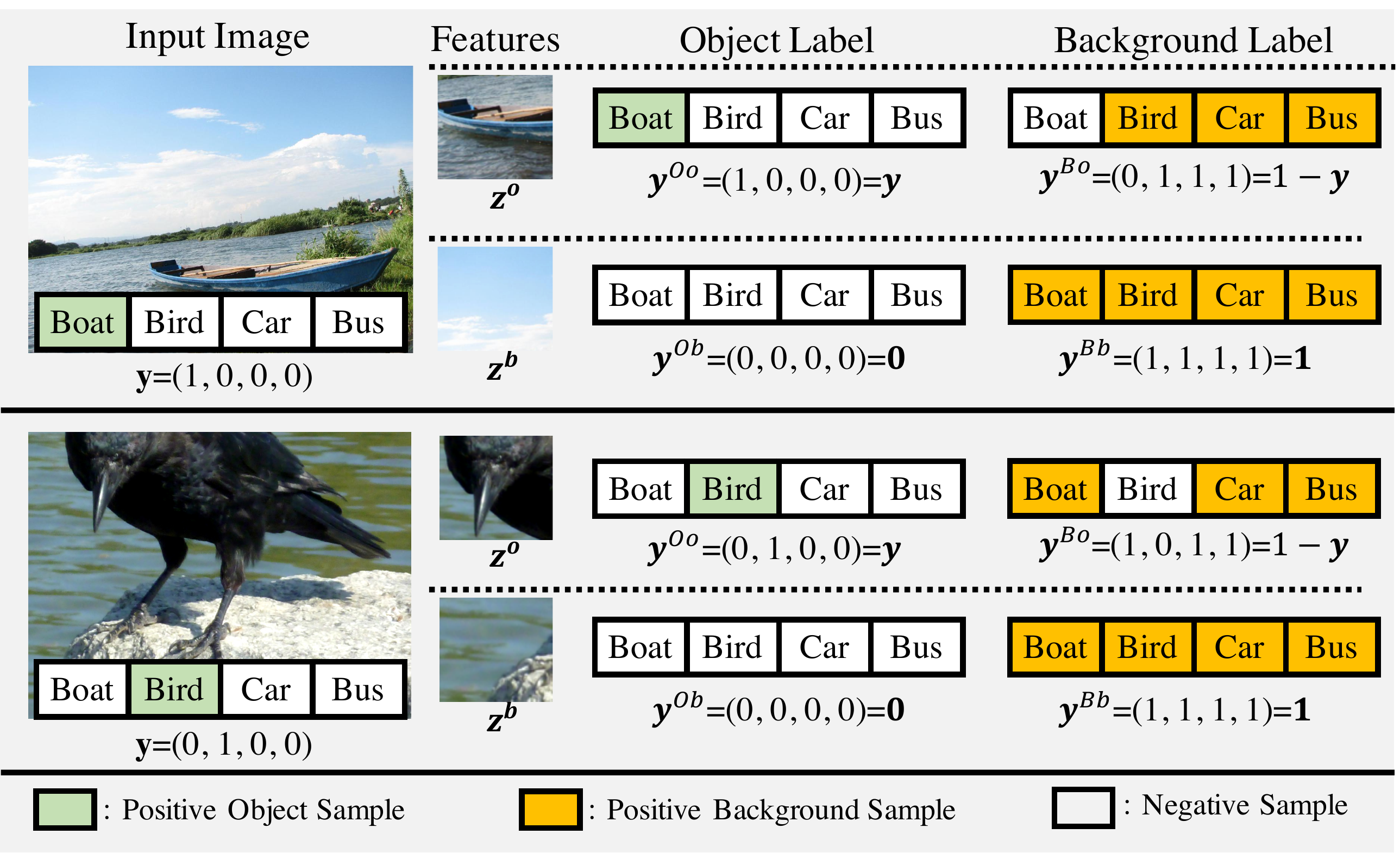}
\caption{Examples for clarifying the properties that forms ground truths.}
\label{fig:loss}
\end{figure}

\subsection{Stagger classification loss}
\label{sec:scloss}

The proposed SC loss is used to train our proposed B-CAM structure based on the image-level mask $\bm{y}$. In general, the SC loss serves as a multi-task loss, which uses $\bm{z}^{o}$ and $\bm{z}^{b}$ as samples to train the object classification and background classification task. The labels of both tasks are collected based on the following properties:

\begin{Assumption}
The feature aggregated by parts of a certain object, i.e. $\bm{z}^{o}$, is the positive sample of this object for the object classification task. Thus, the image-level label $\bm{y}$ is the ground truth of the object classification task for $\bm{z}^{o}$, i.e. $\bm{y}^{Oo} = \bm{y}$.
\end{Assumption}

\noindent For example in Fig.~\ref{fig:loss}, the feature aggregated by the locations of ``the prow of boat'' (or ``head of bird") is the positive object sample for ``boat'' (or ``bird"). 

\begin{Assumption}
The feature aggregated by parts of a certain object, i.e. $\bm{z}^{o}$, is the negative sample of this object for the background classification task. But it is the positive sample of other classes of objects for the background classification task. Thus, $\bm{1} - \bm{y}$ is the ground-truth of the background classification task for $\bm{z}^{o}$, i.e. $\bm{y}^{Bo} = \bm{1} - \bm{y}$, where $\bm{1}$ is a vector filled with 1.
\end{Assumption}
\noindent For example in Fig.~\ref{fig:loss}, the feature aggregated by the locations of ``the prow of boat'' is not the background sample of ``boat'', but it is the background sample of ``bird'', ``car'' and ``bus''. 

\begin{Assumption}
The feature aggregated by some background locations, i.e. $\bm{z}^{b}$, is the negative sample of all objects for object classification. Thus, $\bm{0}$ is the ground truth of the object classification task for $\bm{z}^{b}$, i.e. $\bm{y}^{Ob} = \bm{0}$, where $\bm{0}$ is zero vector.
\end{Assumption}
\noindent  For example in Fig.~\ref{fig:loss}, feature aggregated by the locations of ``water'' or ``sky'' does not belong to any objects, i.e. ``boat'', ``bird'', ``car'' and ``bus''. 

\begin{Assumption}
The feature aggregated by some background locations, i.e. $\bm{z}^{b}$, is the positive sample of all objects for background classification. Thus, $\bm{1}$ is the ground-truth of the background classification task for $\bm{z}^{b}$, i.e. $\bm{y}^{Bb} = \bm{1}$.
\end{Assumption}
\noindent For example in Fig.~\ref{fig:loss}, the feature aggregated by the locations of ``water'' or ``sky'' is the background sample of all objects, i.e. ``boat'', ``bird'', ``car'' and ``bus''.

\begin{algorithm}[h]
\caption{Workflow of the proposed B-CAM} 
\label{alg1} 
\begin{algorithmic}[1] 
\REQUIRE Images set $\mathbb{X}=\{\bm{X}^i\}^{M}_{i=1}$, Labels set $\mathbb{Y}=\{\bm{y}^i\}^{M}_{i=1}$
\WHILE{not reaching stop conditions}
\STATE Calculate the pixel-level features $\bm{Z} \gets e(\bm{X^{i}})$
\STATE Obtain object and background priors $\bm{A}^{o},  \bm{A}^{b}$ by Eq.~\ref{eq:A}
\STATE Generate image-level features $\bm{z}^{o}, \bm{z}^{b}$ by Eq.~\ref{eq:z}
\STATE Extract image-level object classification scores $\bm{s}^{Oo} \gets s^{O}(\bm{z}^{o})$ and $\bm{s}^{Ob} \gets s^{O}(\bm{z}^{b})$ by Eq.~\ref{eq:sse}
\STATE Extract image-level background classification scores $\bm{s}^{Bo} \gets s^{B}(\bm{z}^{o})$ and $\bm{s}^{Bb} \gets s^{B}(\bm{z}^{b})$ by Eq.~\ref{eq:sse}
\STATE Calculate the stagger classification loss $\mathcal{L}_{sc}$ by Eq.~\ref{eq:loss}
%\STATE Backward update the learning parameters 
\STATE \underline{Estimate object localization score $\bm{S}^{O}$ by Eq.~\ref{eq:S}}
\STATE \underline{Estimate background localization score $\bm{S}^{B}$ by Eq.~\ref{eq:S}}
\STATE \underline{Generate the binary localization mask $\bm{Y}^{*}$ by Eq.~\ref{eq:Y}}
\ENDWHILE
\ENSURE  Object localization score $\bm{S}^{O}$, Object classification score $\bm{s}^{Oo}$, Binary localization mask $\bm{Y}^{*}$
\end{algorithmic} 
\end{algorithm}

After obtaining the ground truth of the samples $\bm{z}^{o}$, $\bm{z}^{b}$ on the object and background classification tasks, the SC loss is designed to train our B-CAM with the above sample/label pairs. In detail, the proposed SC loss contains four terms:
\begin{equation}
\begin{split}
\mathcal{L}_{sc} = \lambda_{1}*\mathcal{L}_{O}(\bm{s}^{Oo}, \bm{y}) + \lambda_{2}*\mathcal{L}_{B}(\bm{s}^{Bo}, \bm{1}-\bm{y}) & \\
+ \lambda_{3}*\mathcal{L}_{O}(\bm{s}^{Ob}, \bm{0}) + \lambda_{4}*\mathcal{L}_{B}(\bm{s}^{Bb}, \bm{1})
\end{split}
~,~
\label{eq:loss}
\end{equation}
\noindent where $\mathcal{L}_{O}$ is the object classification loss that is implemented by cross-entropy. $\mathcal{L}_{B}$ is the background classification loss which is implemented as multi-label soft margin loss because a location can be the background of multiple classes. 

In our SC loss, the first term is adopted to supervise the classification accuracy as in other WSOL methods. The third term helps the object classifier perceive pure-background samples and suppress the activation on the background locations of localization maps. Furthermore, the second term and forth term regulate the accuracy of the background scores generated by the background classifier. Thus, if the $\bm{z}^{o}$ and $\bm{z}^{b}$ outputted by the proposed MEA are accurate, the SC loss can train the two estimators of the proposed SSE well.

Moreover, even without using pixel-level supervisions such as the coarse localization maps, our SC loss can still supervise MEA to aggregate features of pure-object and background locations to form $\bm{z}^{o}$ and $\bm{z}^{b}$, respectively. In detail, to show this effect, we take Eq.~\ref{eq:sse} into Eq.~\ref{eq:loss} and split it into two parts:
\begin{equation} 
\left\{
\begin{aligned}
&\mathcal{L}^o_{sc}  = \lambda_{1}*\mathcal{L}_{O}(\mathbf{W}_{O} * \bm{z}^{o}, \bm{y}) + \lambda_{2}*\mathcal{L}_{B}(\mathbf{W}_{B} * \bm{z}^{o}, \bm{1}-\bm{y})\\
&\mathcal{L}^b_{sc}  = \lambda_{3}*\mathcal{L}_{O}(\mathbf{W}_{O} * \bm{z}^{b}, \bm{0}) + \lambda_{4}*\mathcal{L}_{B}(\mathbf{W}_{B} * \bm{z}^{b}, \bm{1})
\end{aligned}
\right. .
\end{equation}

\noindent In this way, $\mathcal{L}^{o}_{sc}$ ensures that $\bm{z}^{o}$ has a high probability of being discerned as a certain object while a low probability of being discerned as the background of other objects. Likewise, $\mathcal{L}^{b}_{sc}$ forces $\bm{z}^{b}$ to be indiscriminating for all classes and have a high probability of being the background of all classes. Thus, aggregating pure-object locations for $\bm{z}^{o}$ and pure-background locations for $\bm{z}^{b}$ will minimize the SC loss.

%%%%%%%%%%%%%%%%

\begin{table*}[!htp]
\caption{Metrics of WSOL methods on CUB-200 dataset}
\centering	
\setlength{\tabcolsep}{3pt}
\begin{tabular}{c|cccc|cccc|cccc|cccc|ccc}
\hline
~ & \multicolumn{8}{c|}{Test Set} & \multicolumn{8}{c|}{Validation Set} &  \multicolumn{3}{c}{~}\\
\hline
~ & \multicolumn{4}{c|}{Top-1} & \multicolumn{4}{c|}{MBA}
& \multicolumn{4}{c|}{Top-1} & \multicolumn{4}{c|}{MBA} 
&  \multicolumn{3}{c}{Complexity}\\
~ & 70\% & 50\% & 30\% & Mean & 70\% & 50\% & 30\% & Mean
& 70\% & 50\% & 30\% & Mean & 70\% & 50\% & 30\% & Mean
& GFlops & MSize & T \\
\hline
CAM & $14.93$ & $54.80$& $70.40$& $46.71$&$19.36$& $72.47$& $94.99$& $62.28$
& 9.40 & 44.10 & 64.30 & 39.27 & 12.80 & 62.30 & 94.20  & 56.43
& $19.13$ & $23.92$ & $\checkmark$\\
\hline
HAS & $22.07$& $51.28$& $64.83$& $46.06$& $30.45$& $70.99$& $92.06$& $64.50$
& 17.20 & 45.90 & 60.40 & 41.17 & 25.50 & 67.50 & 91.10 & 61.37
& $19.13$ & $23.92$  & $\checkmark$\\
ACOL & $12.56$& $42.53$& $54.78$& $36.62$& $19.71$& $70.09$& $93.86$& $61.22$ 
& 8.90 & 33.40 & 48.20 & 30.17 & 14.90 & 58.20 & 89.20 & 54.10
& $63.85$ & $80.55$ & $\checkmark$\\
ADL & $10.72$& $44.30$& $61.63$& $38.89$& $15.05$& $63.31$& $91.47$& $56.61$   
& 6.80 & 34.40 & 55.10 & 32.10 & 10.00 & 53.60 & 89.20 & 50.93
& $19.13$ & $23.92$ & $\checkmark$\\
SPG & $6.42$& $19.12$& $28.56$& $18.04$& $15.26$& $49.93$& $80.19$& $48.46$  
& 4.40 & 15.20 & 23.00 & 14.20 & 12.40 & 44.50 & 73.90 & 43.60
& $56.45$ & $61.67$  & $\checkmark$\\
CutMix & $16.98$& $56.68$& $72.54$ & $48.73$& $20.92$& $71.44$& $93.92$& $62.09$  
& 9.50 & 44.00 & 65.20 & 39.57 & 12.60 & 61.20 & 91.90 & 55.23
& $19.13$ & $23.92$  & $\checkmark$\\
\hline
\textbf{Ours}$^{p}$ & $\textbf{33.91}$& $\textbf{65.31}$& $75.85$& $\textbf{58.36}$& $\textbf{40.42}$& $79.70 $& $95.24 $& $71.79$  
&\textbf{25.40} & \textbf{56.90} & \textbf{72.30} & \textbf{51.53} & \textbf{32.10} & \textbf{73.90} & 95.40 & \textbf{67.13}
& $19.45$ & $24.74$  & $\checkmark$\\
\textbf{Ours}$^{m}$ & $31.41$& $65.46$& $\textbf{76.35}$& $57.74$& $38.21$& $\textbf{81.48}$& $\textbf{96.69}$& $\textbf{72.13}$ 
& 21.50 & \textbf{56.90} & 71.90 & 50.10 & 27.20 & 73.00 & \textbf{95.60} & 65.27
& $19.45$ & $24.74$  & $\times$\\
\hline
\end{tabular}
\begin{flushleft}
$*$ \textit{Text in \textbf{bold style} indicates the best and ``T'' indicates the requirement for threshold searching.}
\end{flushleft}
\label{tab:CUB}
\end{table*}

In summary, Algorithm~1 shows the workflow of the proposed B-CAM, where steps in ``\underline{underline style}'' are only required for the test process. Specifically, during the training process, our B-CAM first calculates the pixel-level feature $\bm{Z}$ by the feature extractor. Then, MEA is utilized to obtain two image-level features $\bm{z}^{o}$ and $\bm{z}^{b}$. Finally, SSE estimates the classification scores used to calculate the SC loss, which subsequently guides the update of learning parameters. During the test process, the pixel-level feature $\bm{Z}$ is directly fed into SSE to generate the binary localization mask $\bm{Y}^{*}$.

\section{Experiments}
\label{sec:experiment}

This section first demonstrates the implementation details of our B-CAM. Then, experiments on different types of datasets are illustrated to validate our proposed B-CAM, including the single-object localization dataset (CUB-200), the noisy single-object localization dataset (OpenImages), and the multi-object localization dataset (VOC2012). Next, ablation studies are presented to confirm the function of different parts of B-CAM. Finally, failed cases and limitations of our B-CAM are discussed to inspire future works.

\subsection{Implementation Details}

The ImageNet pre-trained ResNet~\cite{IMAGENET, RESNET, RESNET38} with the downsample layer of $res4$ and the fully connected layer removed, was used as the feature extractor. All parameters of the two $1 \times 1$ convolution operators $\mathbf{W}_{1}$, $\mathbf{W}_{2}$ were initialized with $0$. While other learnable weight matrices, \textit{i.e.}, $\mathbf{W}_{o}$ and $\mathbf{W}_{b}$ in SSE, were randomly initialized before training. All experiments in this section were conducted with the help of the Pytorch~\cite{pytorch} toolbox on an Intel Core i9 CPU and an Nvidia RTX 3090 GPU. %\textit{Codes are provided in our supplementary material.} 
%The InceptionV3~\cite{INCEPTION} (followed structure of SPG~\cite{SPG}) had been also used for comparison in our experiments. 

Six one-stage WSOL methods were implemented for fair comparison, including CAM~\cite{CAM}, HAS~\cite{HAS}, ACOL~\cite{ACOL}, SPG~\cite{SPG}, ADL~\cite{ADL}, and CutMix~\cite{CUTMIX}. Hyper-parameters of these WSOL methods were set as the optimal settings released by Junsuk~\cite{EVAL}, which were searched 30 trails on the corresponding validation set. For the proposed B-CAM, we evaluated both the object localization score (noted as Ours$^p$), \text{i.e.} $\bm{S}^{O}$, and the final binary mask (noted as Ours$^m$), \text{i.e.} $\bm{Y}^{*}$.

\begin{figure*}
\centering
\includegraphics[width=0.99\textwidth]{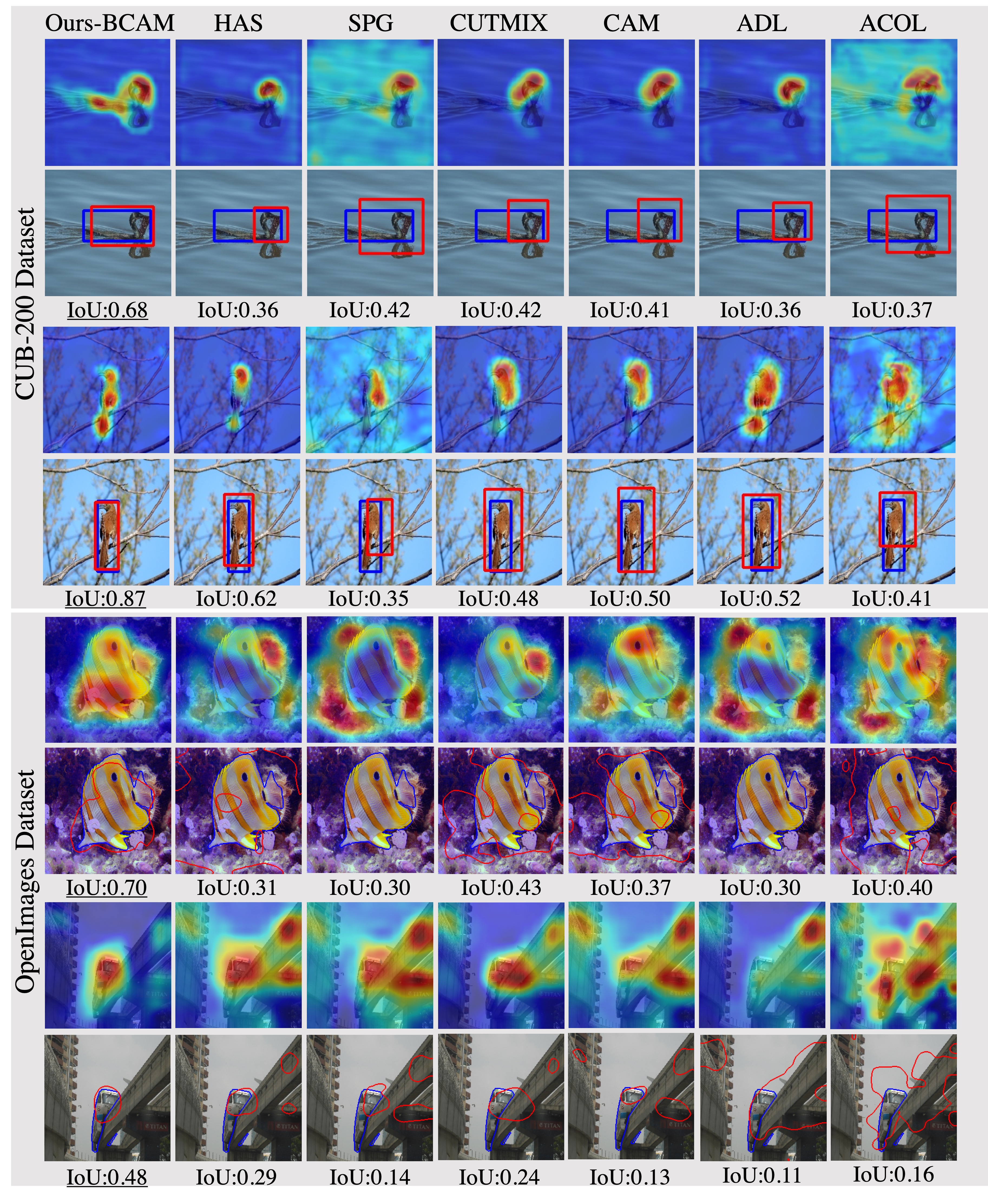}
\caption{Visualizations of the object localization scores and predicted bounding boxes of WSOL methods on the CUB-200 and OpenImage datasets. The ground truth bounding boxes/object boundaries are noted in blue color, while the predicted bounding boxes/object boundaries are noted in red. Note that the bounding boxes and localization masks with the highest IoU among all thresholds are visualized for each method in these figures.}
\label{fig:cub_val_vis}
\end{figure*}

\subsection{Experiments on Single Object Localization}
\subsubsection{Setting}
The experiments of single object localization were conducted on the CUB-200 dataset~\cite{CUB}. It contains $11,788$ \textit{single-class images} annotated for $200$ classes with the corresponding \textit{object bounding box annotations} to benchmark the localization tasks. Following the official setting, $5,994$ images were used as the training set to train the WSOL methods with only image-level labels and the other $5,794$ images were used to report the performance. Additionally, $1,000$ extra images (5 images per class) annotated by Junsuk~\cite{EVAL} were adopted as the validation set to search the optimal hyper-parameters.

In the training process, the input images were resized to $256\times256$, and then randomly cropped to $224\times224$ followed by a random horizontal flip operation to form the batches of $32$ images. Hyper-parameters were set as $M=60$, $\lambda_1=1$, $\lambda_2=\lambda_3=0.3$, and $\lambda_4=0.2$. The SGD optimizer with weight decay $1e$-$4$, and momentum $0.9$ was used to train the WSOL models for $50$ epochs.The initial learning rate was set as $1.7e$-$4$, which was divided by $10$ every $15$ epochs. %Note that, the learning rate and other method-specific hyper-parameters of other compared WSOL methods are set as the optimal settings released by Junsuk~\cite{EVAL}, which were searched 30 trails on the validation set.

The maximal box accuracy (MBA)~\cite{EVAL} was used to evaluate the bounding boxes generated by the localization map. Specifically, for each background threshold $\tau$, the largest connected component of the corresponding binary mask was used as the predicted bounding box. Then, the box accuracy under certain $\tau$ was calculated by counting the number of images where the intersection over union (IoU) between the predicted box and the ground truth box was higher than a ratio $\delta$. The maximum scores for all possible thresholds were reported as MBA. Moreover, we also used the top-1 localization accuracy (Top-1) to evaluate both localization and classification of the WSOL methods.

%%%%%
\subsubsection{Results}

Table.~\ref{tab:CUB} shows the performance of our B-CAM and the other six one-stage methods on the CUB-200 test set that is challenging for one-stage WSOL methods because only bird images are present. Our proposed B-CAM significantly improves the quality of the object localization map (Ours$^{p}$) and achieves better performance on nearly all evaluation metrics for this fine-grained dataset ($7.63\%$ higher in MBA, $9.63\%$ higher in Top-1 than the best of others) with only a minor complexity increase (0.3 GFlops). This excellent improvement benefits from the trait that our B-CAM can perceive the unseen pure-background samples (images without birds) by the background feature $\bm{z}^b$ and use it to suppress the localization score of the background area. Moreover, the background localization map $\bm{S}^{B}$ of our B-CAM can also release the background threshold searching process. Directly adopting the background score map $\bm{S}^{B}$ as the binary map (Ours$^{m}$) only causes a little reduction in metrics of $70\%$ IoU.

Moreover, we also plotted the performance of WSOL methods under different thresholds in Fig.~\ref{fig:intro_perform}. It can be seen that the peak value of our localization map is the highest among all the WSOL methods, which shows the effectiveness of our SC loss to reduce the activation of background location. Though using the adaptive background score generated by our background classifier will lower the peak performance, it releases the post threshold searching step, which influences the performance of one-stage WSOL methods.

We also compared our B-CAM with some other state-of-the-art WSOL methods on the CUB-200 dataset in Table.~\ref{tab:CUB_SOTA} with the GT-Known/Top-1 localization metrics reported by their papers. Note that some of these methods used the optimal hyper-parameter on the test set rather than following the fair WSOL evaluation criterion~\cite{EVAL} to search them on the validation set. Even so, our B-CAM still shows the best performance with the highest GT-Known metric among all methods.  Only the Top-1 metric is a bit lower than the multi-stage method PSOL~\cite{PSOL}, which adopts two additional stages and only generate class-agnostic localization bounding boxes rather than the class-specific localization mask as our B-CAM.

\begin{table}%[!htp]
\caption{Comparing with SOTA methods on the CUB-200 test set}
\centering	
\setlength{\tabcolsep}{3pt}
\begin{tabular}{c|c|cc|c}
\hline
~ & Extractor & Top-1 Loc & GT-Known & Stages \\
\hline
ADL~\cite{ADL} & InceptionV3 & 53.04 & - 
& 1 \\
DANet~\cite{DANet} & InceptionV3 & 49.45 & 67.03 
& 1 \\
I2C~\cite{I2C} & InceptionV3 & 55.99 & - 
& 1 \\
GCNet~\cite{GC} & InceptionV3 & 58.58 & 75.30 
& 3 \\
MEIL~\cite{MEIL} & InceptionV3 & 57.46 & - 
& 1 \\
UPSP~\cite{UPSP} & InceptionV3 & 53.59 & 72.14
& 2 \\
DGL~\cite{DGL} & InceptionV3 & 51.50 & 66.64 
& 1\\
SLT~\cite{SLT} & InceptionV3 & 55.70 & 67.70 
& 3 \\
\hline
ADL~\cite{ADL} & ResNet50-SE & 62.29 & - 
& 1 \\
R-CAM~\cite{RCAM} & ResNet50-SE & 58.39 & 74.51 
& 1 \\
R-ADL~\cite{RCAM} & ResNet50 & 59.53 & 77.58 
& 1 \\
R-HAS~\cite{RCAM} & ResNet50 & 57.42 & 75.34 
& 1 \\
Mixup~\cite{CUTMIX} & ResNet50 & 49.30 & - 
& 1 \\
Cutout~\cite{CUTMIX} & ResNet50 & 52.78 & - 
& 1 \\
Cutmix~\cite{CUTMIX} & ResNet50 & 54.81 & - 
& 1 \\
ICLCA~\cite{ICLCA} & ResNet50 & 56.10 & 72.79 
& 1 \\
PSOL~\cite{PSOL} & ResNet50 & \underline{\textbf{68.17}} & - 
& 3 \\
CAAM~\cite{CAAM} & ResNet50 & 62.58 & 75.22 
& 1 \\
CAAM-Mix~\cite{CAAM} & ResNet50 & 64.70 & 77.35 
& 1 \\
DGL~\cite{DGL} & ResNet50 & 61.72 & - 
& 1 \\
\hline
\textbf{Ours}$^p$ & ResNet50 & 65.32 & 79.70 
& 1\\
\textbf{Ours}$^m$ &ResNet50 & \textbf{65.46} & \underline{\textbf{81.48}}
& 1\\
\hline
\end{tabular}
\begin{flushleft}
$*$ \textit{Text in \underline{\textbf{underline bold style}} indicates the best among \textbf{all methods}.} \\
$*$ \textit{Text in \textbf{only bold style} indicates the best among \textbf{one-stage method}.}
%$*$ \textit{Text in \underline{underline style} indicates the multi-stage WSOL method.}
\end{flushleft}
\label{tab:CUB_SOTA}
\end{table}

%\begin{figure}
%\centering
%\includegraphics[width=0.49\textwidth]{pics/cub_open_vis.png}
%\caption{Visualizations of the object localization scores and predicted bounding boxes of WSOL methods on the CUB-200 and OpenImage datasets. The ground truth bounding boxes/object boundaries are noted in blue color, while the predicted bounding boxes/object boundaries are noted in red. Note that the bounding boxes and localization masks with the highest IoU among all thresholds are visualized for each WSOL method in these figures.}
%\label{fig:cub_val_vis}
%\end{figure}

To qualitatively represent the performance of the WSOL methods, the localization map and bounding boxes with optimal threshold are visualized in Fig.~\ref{fig:cub_val_vis}. It can be seen that SPG~\cite{SPG} and ACOL~\cite{ACOL} seriously suffer from the excessive activation of the background locations, especially for the objects with object-related background (duck/water in the first two rows). This is because these two methods both affirm the locations with high activation (may contain object-related background) belong to the object parts. Though the methods that adopt random-erasing augmentation (HAS~\cite{HAS}, ADL~\cite{ADL}, CUTMIX~\cite{CUTMIX}) can better catch object parts than CAM~\cite{CAM}, they cannot effectively suppress the activation of the background locations, especially near object boundaries. This makes the localization map generated by these methods still larger than the real objects. Compared with those methods, our B-CAM can activate more object parts and avoid excessive background activation, which is benefitted from our awareness of background cues. Thus, the localization boxes generated by our B-CAM have higher IoU than others.

%%%%%%%%%%%%%%%%%%%%%%%%%%%%%%%%%%%%%%%%

%\subsection{Experiments on OpenImages Dataset}
\subsection{Experiments on Noisy Single Object Localization}

%%%%%
\subsubsection{Setting}

Experiments were also conducted on the OpenImages WSOL dataset, which contains some noisy image labels. This dataset~\cite{OPENIMAGE, EVAL} has 3,7319 images of 100 classes, where 2,9819, 2,500, and 5,000 images serve as the training set, validation set, and test set, respectively. Unlike the CUB-200 dataset, the OpenImages dataset provides \textit{pixel-level object binary masks} with the \textit{single-class image annotation} for each image. The annotations of this dataset contain noise, \textit{i.e.} only \textit{the most conspicuous object} is annotated in images with multiple objects. For example, an image containing both ``boat'' and ``bird'' is only annotated as ``boat''.

In the training process, $M=80$, $\lambda_1=\lambda_3=\lambda_4=1$ and $\lambda_2=0.5$ were used. Settings of data pre-processing and SGD were the same as the CUB-200 dataset. Each method was trained 10 epochs and the learning rate was divided by $10$ every $3$ epochs. $1.7e$-$4$ was adopted as the initial learning rate for our B-CAM. Note that the learning rate and other method-specific hyper-parameters of other WSOL methods were also set as the optimal settings released by Junsuk~\cite{EVAL}.

%%%%%
\subsubsection{Results}
\begin{figure}%[!htp]
\centering
\subfigure[IOU on OpenImages]{
\includegraphics[width=0.225\textwidth]{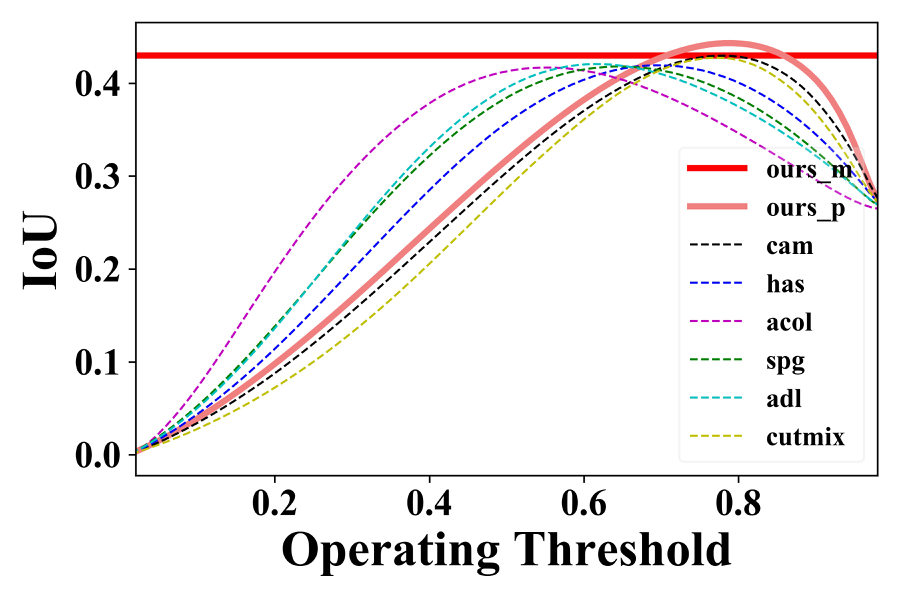}
\label{fig:miou_open}}
\subfigure[P-R Curve on OpenImages]{
\includegraphics[width=0.225\textwidth]{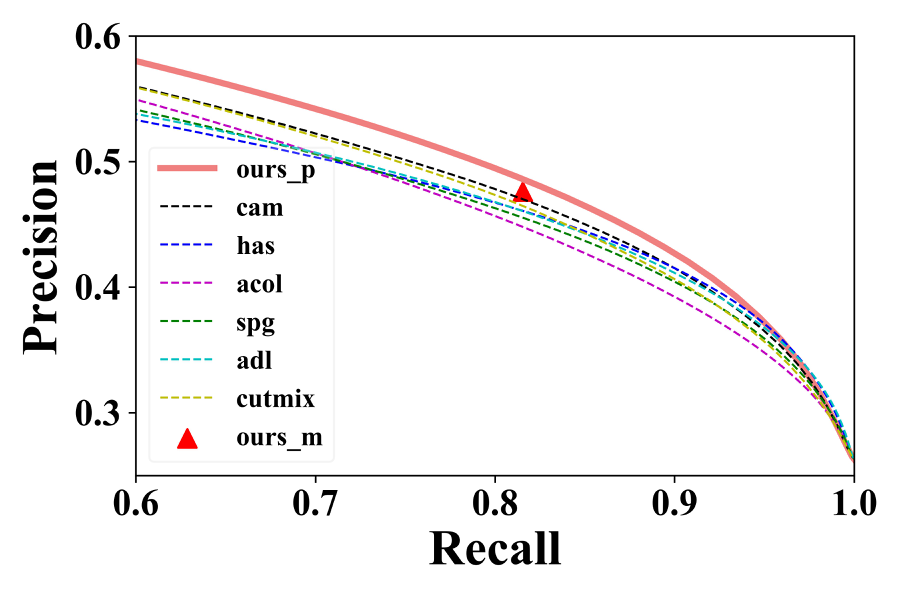}
\label{fig:pr_open}}
\caption{Threshold-related metrics on OpenImages dataset. Metrics of our B-CAM are highlighted with solid lines. (a) IoU with different thresholds. (b) P-R curve plotted with different thresholds. }
\label{fig:exps}
\end{figure}

The IoU between the ground truth and predicted binary mask was used to quantitatively evaluate the WSOL methods, where the predicted binary mask can be obtained by thresholding the localization map generated by the WSOL methods with parameter $\tau \in (0,1)$. Corresponding results of different background thresholds are shown in Fig.~\ref{fig:miou_open}. It shows that the results are in accordance with the CUB-200 dataset, and the peak of our localization map (Ours$^{p}$) is the highest among all the WSOL methods. Though our binary mask  (Ours$^{m}$) has a relatively lower peak than our localization map (Ours$^{p}$), it is still higher than all other WSOL methods and avoids the post-threshold searching step. Moreover, the precision-recall (P-R) curves of the localization maps were also plotted based on the precision/recall pairs of different background thresholding scales for evaluation. The P-R curve of our B-CAM is located closer to the top right corner in Fig.~\ref{fig:pr_open}, which also indicates our effectiveness. 

We also used peak intersection over union (pIoU)~\cite{SEM} under all background thresholds and the threshold-free pixel average precision (PxAP) metrics~\cite{EVAL} to quantitatively evaluate object localization performance on the OpenImages dataset. Table.~\ref{tab:Open_val_test} shows the pIoU and PxAP metrics of all WSOL methods and our proposed B-CAM on OpenImages validation set and test set. Our method obtains the maximal improvement over the original CAM (1.36 higher pIoU and 1.27 higher PxAP on test set) among all WSOL methods. Note that we cannot calculate the PxAP (area under the P-R curve) of our binary masks whose P-R curve degrades into a dot because of its insensitivity to the thresholds. 
\begin{table}%[!htp]
\caption{Mask-based Metrics on OpenImage and CUB-200 dataset}
\centering
\begin{tabular}{c|cc|cc|cc}
\hline
~ & \multicolumn{4}{c|}{OpenImage (Noise)} & \multicolumn{2}{c}{CUB-200 (Clean)}\\
\hline
~ & \multicolumn{2}{c|}{Test Set} & \multicolumn{2}{c|}{Validation Set} & \multicolumn{2}{c}{Test Set}\\
~ & pIoU & PxAP & pIoU & PxAP & pIoU & PxAP \\
\hline
CAM & 42.95 & 58.19
& 43.42 & 58.59  
& 47.60 & 66.78 \\
HAS & 41.92 & 55.10
&  42.47 & 55.84 
& 49.82 & 71.32 \\
ACOL & 41.68 & 56.37
& 42.73 & 57.70  
& 41.56 & 56.78 \\
ADL & 42.05 & 55.02
& 42.33 & 55.26  
& 42.29 & 56.96 \\
SPG & 41.79 & 55.76
& 42.17 & 56.45 
& 31.71 & 48.45 \\
CutMix & 42.73 & 57.47
& 43.43 & 58.18 
& 45.89 & 64.64 \\
\hline
\textbf{Ours}$^{p}$ &\textbf{44.31} & \textbf{59.46}
& \textbf{44.73} & \textbf{60.27} 
&\textbf{57.14} & \textbf{78.76} \\
\textbf{Ours}$^{m}$ & 42.98 & -
& 43.70 & -
& 53.31 & - \\
\hline
\end{tabular}
\label{tab:Open_val_test}
\end{table}

%%%
As a comparison to the single-object localization with clean labels, we also used the recently released localization mask on the CUB-200 test set to evaluate the pIoU~\cite{SEM} and PxAP~\cite{EVAL} for the predicted localization masks. Corresponding results are given in Table.~\ref{tab:Open_val_test}. Compared with the dataset with clean labels, the noisy labels will restrain the performance of our B-CAM ($1.44$ higher on noise dataset and $9.54$ higher on clean dataset than CAM). This is because the noisy label violates the four properties used to design our SC loss discussed in Sec.~\ref{sec:scloss}. Considering the case above (an image containing both ``boat'' and ``bird''), the feature aggregated by the background locations of ``boat'' may contain parts of ``bird'', i.e. may not be the background of ``bird'', which violets the Property 2 and weakens the localization performance.

Finally, localization binary masks generated by the WSOL methods on the OpenImages dataset are also visualized in Fig.~\ref{fig:cub_val_vis}. Though the unlabeled object classes in OpenImages dataset may weaken the effect of our B-CAM, the localization mask of our B-CAM can still cover more object locations and achieve the best IoU metric among all WSOL methods. In addition, the localization mask of our B-CAM also has better boundary adherence and avoids containing object-related background locations (such as the track of train) due to our awareness of background cues.

\begin{figure*}[!htp]
\centering
\includegraphics[width=0.99\textwidth]{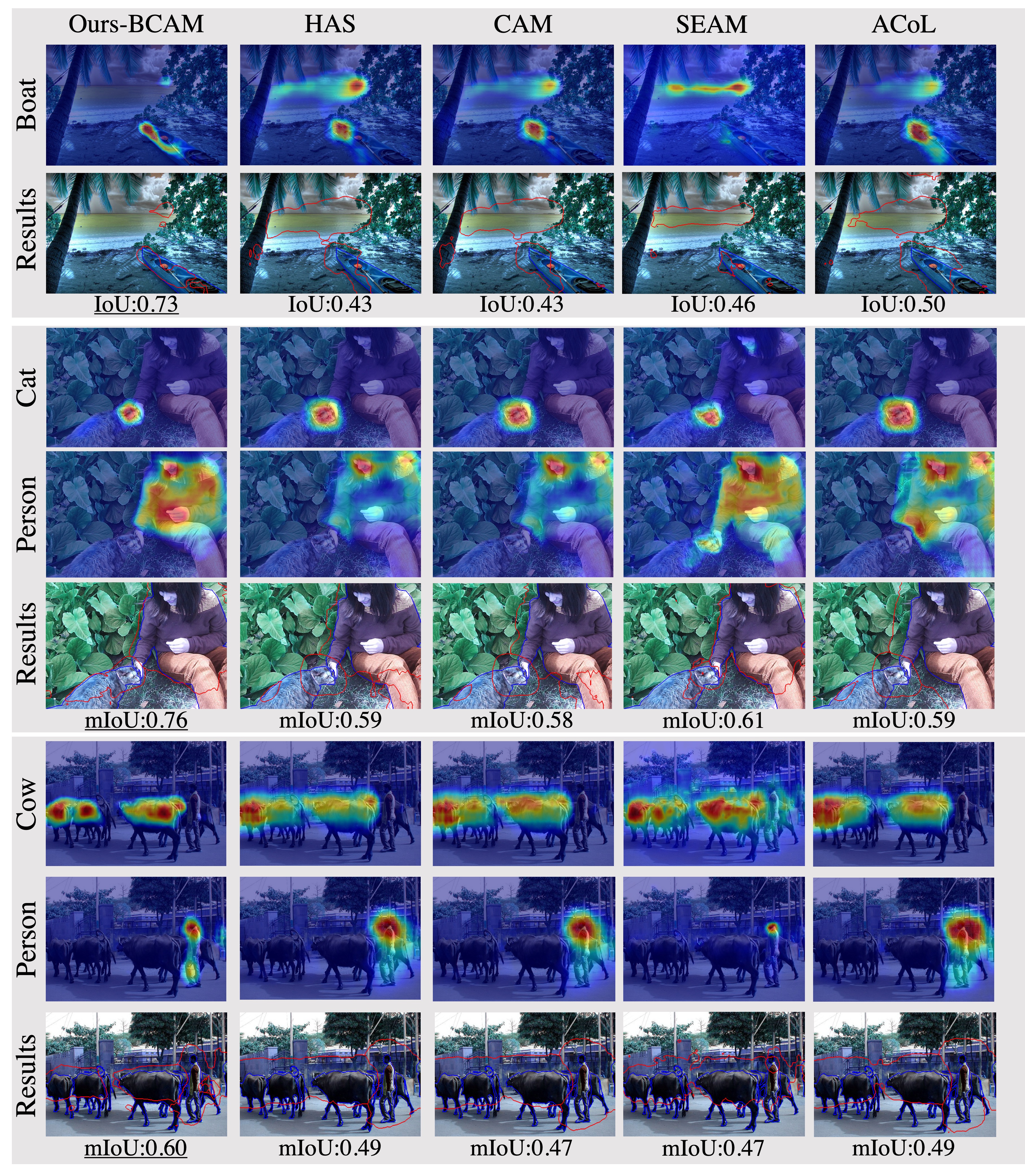}
\caption{Visualizations of the localization scores of WSOL methods on the VOC2012 dataset. The ground truth object boundaries are noted in blue color, while the predicted bounding object boundaries are noted in red.}
\label{fig:voc_vis}
\end{figure*}

\subsection{Experiments on Multi-object Localization}
\subsubsection{Setting}
The multi-object localization dataset VOC2012 was also used to evaluate the proposed B-CAM, where all the objects with different classes are annotated for a certain image. The VOC2012 dataset~\cite{VOC} contains 14,978 images of 20 classes, where $10,582$ images are annotated by SBD~\cite{SBD}. These $10,582$ images (including the $1,464$ images from the training set of official dataset separation) are used as the train set for our experiments. The official training set and validation set  contain $1,464$ and $1,449$ images, respectively, which are used to search hyper-parameters and report results. Unlike the CUB-200 and OpenImages datasets, the annotation of the VOC2012 dataset gives the \textit{multi-class image annotation} with \textit{pixel-level object binary mask}, \textit{i.e.}, annotating all the objects that exist in an image.

ResNet38~\cite{RESNET38} was used as the feature extractor for this dataset to guarantee fair comparison with the existing method~\cite{SEAM}. Considering that VOC2012 is a multi-class object localization dataset, we adopted multi-label soft margin loss as the $\mathcal{L}_{O}$ to replace the cross-entropy when reproducing the WSOL methods. In the training process, the input images were first randomly resized into range (448, 768), and then cropped into $448\times448$ followed by a color jittering operation to form batches of $8$ images. The hyper-parameters were set as $M=20$ and $\lambda_1=\lambda_2=\lambda_3=\lambda_4=1$. SGD optimizer with weight decay $1e$-$5$ and momentum $0.9$ was used to train the WSOL models for a total of $8$ epochs. The initial learning rate was set as $0.01$, which was delayed by the poly strategy. 

\subsubsection{Results}

The pIoU metric and its corresponding sensitivity (SE), precision (PR), and specificity (SP) were used to evaluate the performance. Results of our B-CAM and other WSOL methods including CAM~\cite{CAM}, ACOL~\cite{ACOL}, HAS~\cite{HAS}, and SEAM~\cite{SEAM} are shown in Table.~\ref{tab:VOC}. It shows that those object localization methods cannot effectively improve the original CAM on the VOC2012 dataset that contains multi-objects in an image. However, our B-CAM can improve the performance to a great extent (7.89\% and 7.16\% higher mIoU in validation set and test set), owing to our background awareness. The background classifier of our B-CAM can be better trained than with the OpenImages dataset, because every object exists in the image is annotated by the VOC2012 dataset. Moreover, compared with the class-agnostic post-thresholding used by other WSOL methods, our background classifier can also generate the background score for each class, which is more reasonable for multi-object localization. So our binary masks (Ours$^m$) even have a higher mIoU than localization scores (Ours$^p$).
\begin{table}%[!htp]
\centering	
\setlength{\tabcolsep}{3.5pt}
\caption{Metric of WSOL methods on VOC2012 dataset}
\begin{tabular}{c|cccc|cccc}
\hline
~ & \multicolumn{4}{c|}{Official Train Set} & \multicolumn{4}{c}{Official Validation Set} \\
~ & pIoU & SE & PR & SP & pIoU & SE & PR & SP \\
\hline
CAM & 45.43 & 43.53 & 55.64 & 33.77 
& 46.60 & 43.67 & 56.30 & 33.15 \\
HAS & 45.14 & 43.50 & 55.32 & 33.79  
& 46.32 & 43.72 & 56.02 & 33.26 \\
ACOL & 45.28 & 42.71 & 55.51 & 33.97  
& 46.60 & 42.92 & 56.08 & 32.57 \\
SEAM & 49.68 & 51.09 & 62.81 & 41.13 
& 51.78 & 52.01 & 64.10 & 40.86 \\
\hline
\textbf{Ours}$^{p}$ & 52.64 & 56.08 & 69.52 & 50.75
 & 54.43 & 56.26 & 70.09 & 50.51 \\
\textbf{Ours}$^{m}$ & \textbf{52.69} & \textbf{56.18} & \textbf{69.91} & \textbf{51.17} 
& \textbf{54.51} & \textbf{56.38} & \textbf{70.49} & \textbf{50.96} \\
\hline
\end{tabular}
\label{tab:VOC}
\end{table}

\begin{table}%[!htp]
\caption{The mIoU of each classes on VOC2012 official validation dataset}
\centering	
\setlength{\tabcolsep}{1pt}
\begin{tabular}{c|ccccccccccc}
\hline
~ & bg & \underline{plane} & bike & \underline{bird} & \underline{boat} & bottle & bus & car & cat & chair  & cow  \\
\hline
CAM & 73.0 & 35.7 & 24.0 & 40.1 & 26.6 & 41.7 & 64.4 & 53.0 & 52.2 & 24.6 & 48.5 \\
HAS & 73.0 & 35.7 & 24.0 & 39.2 & 25.6 & 41.4 & 63.8 & 52.9 & 53.1 & 24.3 & 48.2\\
ACOL & 72.0 & 33.7 & 23.9 & 38.4 & 25.6 & 45.4 & \textbf{67.4} & 54.3 & 52.1 & 23.4 & 48.9 \\
SEAM & 80.0 & 47.4 & 25.9 & 46.3 & 31.4 & 48.0 & 53.5 & 59.0 & 55.3 & 26.8 & 49.5\\
\textbf{Ours}$^p$ & 82.4 & 54.0 & 29.2 & 54.7 & 39.2 & 48.4 & 59.1 & 59.1 & \textbf{69.1} & 30.5 & \textbf{50.0} \\
\textbf{Ours}$^m$ & \textbf{82.5} & \textbf{54.6} & \textbf{29.2} & \textbf{55.1} & \textbf{39.4} & 48.2 & 59.4 & \textbf{59.3} & 69.1 & \textbf{30.6} & 49.1 \\
\hline
~ & table & dog & horse & motor & man & plant & sheep & sofa & \underline{train} & tv & \textbf{avg}\\
\hline
CAM & 44.1 & 53.2 & 49.1 & 56.4 & 49.6 & 32.8 & 53.5 & 46.0 & 48.6 & 37.0 & 46.6\\
HAS & 43.5 & 53.3 & 48.6 & 56.4 & 50.5 & 32.8 & 53.3 & 45.7 & 48.9 & 33.6 & 46.3\\  
ACOL & 43.9 & 52.3 & 48.7 & 57.1 & 46.9 & 33.0 & 53.0 & \textbf{46.7} & 45.4 & 39.1 & 46.6 \\
SEAM & \textbf{45.9} & 58.3 & 51.0 & 58.1 & 58.8 & 40.0 & 63.0 & 50.3 & 54.3 & \textbf{40.7} & 51.8 \\
\textbf{Ours}$^p$ & 36.3 & 71.2 & 57.0 & 59.9 & \textbf{64.1} & \textbf{40.8} & 60.6 & 42.9 & 60.9 & 37.1 & 54.4 \\
\textbf{Ours}$^m$ & 36.3 & \textbf{71.4} & \textbf{56.1} & \textbf{59.9} & 64.1 & 40.7 & \textbf{60.6} & 43.1 & \textbf{61.0} & 36.9 & \textbf{54.5} \\
\hline
\end{tabular}
\label{tab:VOC_class}
\end{table}

We also exhibit the performance of the 20 classes on the VOC2012 dataset in Table.~\ref{tab:VOC_class}, where our B-CAM obtains better performance nearly on all the categories, especially for the categories with an object-related background (20.90\% IoU higher for ``plane'', 14.4\% higher IoU for ``train'' and 13.8\% higher for ``boat''). Moreover, for the background class, our B-CAM also has a much larger improvement (10.56 \% higher IoU), which shows the effectiveness of our B-CAM for suppressing the background activation.

Finally, the localization maps of those methods are visualized in Fig.~\ref{fig:voc_vis}, where the masks are selected by the ones with the highest mIoU among all background thresholds. It shows that all other methods face excessive activation on the background locations, especially the object-related background (water locations for the boat image). Moreover, when facing images with multi-objects, the localization maps of SEAM are also contaminated by those classes. For example, the locations of cat/person (the second/third images) also have high activation on the localization map of person/cow. However, our B-CAM can perceive the background cues of each class, which can avoids this problem and improves our localization maps.

%\begin{figure}[!htp]
%\centering
%\includegraphics[width=0.49\textwidth]{pics/voc_vis.png}
%\caption{Visualizations of the localization scores of WSOL methods on the VOC2012 dataset. The ground truth object boundaries are noted in blue color, while the predicted bounding object boundaries are noted in red.}
%\label{fig:voc_vis}
%\end{figure}

\subsection{Ablation Studies}
\label{sec:ablation}
Ablation studies were also conducted for our proposed B-CAM. We first explored the effectiveness of all the proposed parts of our method by conducting three types of B-CAM: 1) Ours$_1$ only used our object aggregator (OA) to replace the original GAP-based aggregator of CAM; 2) Ours$_2$ further added the background aggregator (BA) that helps to train an additional background classifier (BC); 3) Ours$_3$ used the complete SSE that added the stagger path (SP) for generating $\bm{s}_{ob}$ upon Ours$_2$ to suppress the background activation. All models contained the object classifier and adopted the same initialization weights for the common parts. 
\begin{table}%[!htp]
\caption{Ablation studies on the CUB-200 test set}
\center
\setlength{\tabcolsep}{3pt}
\begin{tabular}{c|cc|cccc|c}
\hline
~ & Top-1 Mean & MBA Mean& OA& BA & BC& SP & T\\
\hline
CAM & $46.71$ & 62.28
& $\times$ & $\times$ & $\times$ & $\times$ & $\checkmark$\\
\hline
Ours$_{1}$ & $46.40$ & 56.06
& $\checkmark$ & $\times$ & $\times$ & $\times$ & $\checkmark$\\
Ours$_{2}$ & $49.67$ & 59.75 
& $\checkmark$ & $\checkmark$ & $\checkmark$ & $\times$ & $\checkmark$\\
\hline
Ours$_{3}^{p}$ & $58.43$ & 72.28
& $\checkmark$ & $\checkmark$ & $\checkmark$ & $\checkmark$ & $\checkmark$\\
Ours$_{3}^{m}$ & $57.25$ & 71.54
& $\checkmark$ & $\checkmark$ & $\checkmark$ & $\checkmark$ & $\times$\\
\hline
\end{tabular}
\label{tab:ablation}
\end{table}

Table.~\ref{tab:ablation} shows the results of these B-CAMs. It illustrates that instead of enhancing the performance, only using OA (Ours$_1$) even drops the performance compared with the baseline. This is because in such a condition, the  object feature is only coarsely formed by OA without any restrictions, which may undesirably contain excessive background or missing object parts. When adding BA and BC (Ours$_2$), additional restrictions can be added based on $\mathcal{L}^b_{sc}$ to ensure that the object feature is not classified into the background, which enhances the purity of the object feature. Thus the quality of our localization map raises about $3.27\%$ in Top-1. Next, when adopting the complete SSE, $\bm{s}^{Ob}$ can help to suppress the background activation on the localization map (Ours$^p_3$) by the second term of $\mathcal{L}^o_{sc}$, which brings an $8.76\%$ improvement over Ours$_2$. Finally, when directly evaluating the binary mask (Ours$^m_3$), the supervised thresholding can be removed with only a $1.25\%$ drop in Top-1.

%Experiments were also conducted to illustrate the effect of adopting different numbers of spatial attention maps. We found that the influence of hyper-parameter $M$ on the performance is not monotonic.  As shown in Fig.~\ref{fig:ablation}, it can be seen that the performance first increases when $M<60$, because the attention maps are insufficient to catch all the possible spatial relations for different objects/stuff. Then, when $M \in (60, 140)$, the performance tends to level off, which means the attention is relatively suitable. Finally, when $M>140$, the performance faces a dramatic decrease. A possible explanation is that when $M$ is too large, the redundancy attention maps will catch the most discriminative localization, which lowers the effect of other object parts on the final object feature. Thus, the performance will nearly drop to a similar value as $M=1$. 
%
%\begin{figure}[!htp]
%\center
%\subfigure{
%\includegraphics[width=0.23\textwidth]{pics/ablation_1.png}}
%\subfigure{
%\includegraphics[width=0.23\textwidth]{pics/ablation_2.png}}
%\caption{Metrics when using different hyper-parameter $M$ for B-CAM.}
%\label{fig:ablation}
%\end{figure}

To verify our background classifier, we evaluated our background localization score on the CUB-200 and OpenImages datasets. Specifically, different thresholds are adopted to the background localization score to generate the background localization mask. Then, for an image with class $k$, we use $1-\bm{Y}_{k, :}$ as the ground truth of the background localization task to calculate the pIoU and PxAP metrics that evaluate our background localization score. Corresponding scores are given in Table.~\ref{tab:eval_back}, where the background localization maps of our B-CAM obtain satisfactory scores on these datasets. This indicates the effectiveness of our background classifier.  
\begin{table}%[!htp]
\caption{Metrics of the background localization score}
\centering	
\begin{tabular}{c|cc}
\hline
~ & pIoU & PxAP \\
\hline
OpenImage Validation Set & 72.75 & 69.28 \\
OpenImage Test Set & 73.71 & 69.95 \\
CUB-200 Test Set & 86.66 & 81.71 \\
\hline
\end{tabular}
\label{tab:eval_back}
\end{table}

To confirm that our better localization map is not attributed to calibration dependency~\cite{EVAL}, we also explored the upper bound performance for our B-CAM and other WSOL methods. Specifically, we searched the optimal image-scale (OIS) threshold to generate the binary mask based on the localization map for evaluation. Table.~\ref{tab:CUB_OIS} shows the scores of our B-CAM and other one-stage WSOL methods. It can be seen owing to suppressing the activation localization, our B-CAM still outperforms other methods to a great extent. This guarantees the effectiveness of our B-CAM in improving the upper bound quality of the localization map.
\begin{table}%[!htp]
\caption{The metrics in OIS scale of WSOL methods on CUB-200 test set}
\centering	
\setlength{\tabcolsep}{3pt}
\begin{tabular}{c|cccc|cccc}
\hline
~ & \multicolumn{4}{c|}{Top-1 OIS} & \multicolumn{4}{c}{MBA OIS} \\
~ & 70\% & 50\% & 30\% & Mean & 70\% & 50\% & 30\% & Mean \\
\hline
CAM & 34.00 & 69.62 & 72.99 & 58.87 & 45.36 & 94.11 & 99.76 & 79.74 \\
\hline
HAS & 42.94 & 65.72 & 68.69 & 59.12 & 60.68 & 94.46 & 99.65 & 84.93 \\
ACOL & 28.51 & 53.37 & 56.92 & 46.27 & 45.84 & 91.27 & 98.84 & 78.65 \\
ADL & 25.63 & 59.98 & 64.77 & 50.13 & 37.61 & 90.40 & 99.67 & 75.89 \\
SPG & 37.33 & 72.16 & 75.73 & 61.74 & 47.58 & 94.11 & 99.78 & 80.49 \\
CutMix & 16.52 & 28.86 & 32.65 & 26.01 & 43.08 & 83.59 & 97.10 & 74.59 \\
\hline
Ours & \textbf{57.90} & \textbf{76.94} & \textbf{78.13} & \textbf{70.99} & \textbf{72.39} & \textbf{97.88} & \textbf{99.97} & \textbf{90.08} \\
\hline
\end{tabular}
\begin{flushleft}
\end{flushleft}
\label{tab:CUB_OIS}
\end{table}

\begin{figure*}%[!htp]
\centering
\includegraphics[width=0.99\textwidth]{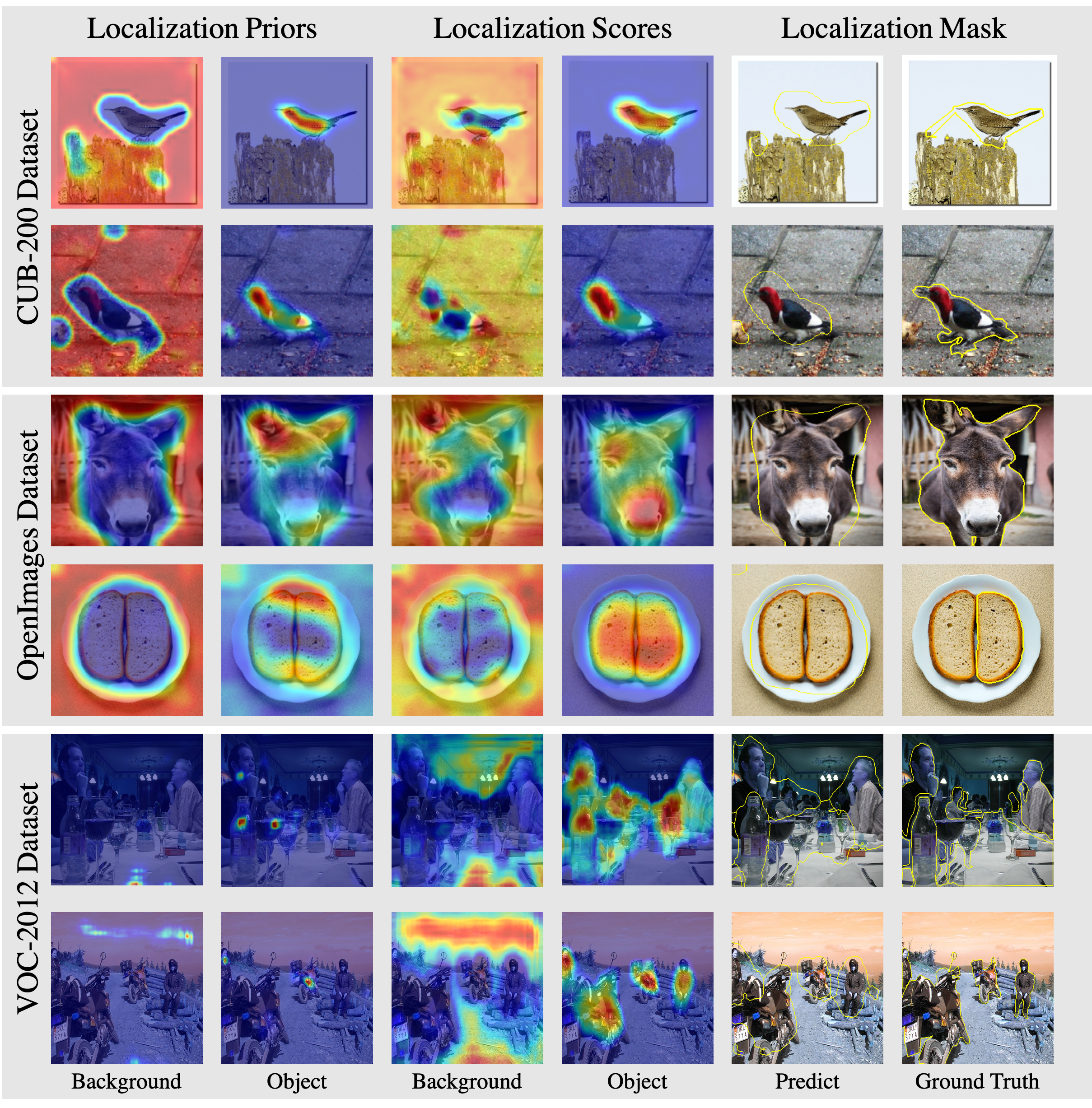}
\caption{Visualizations for the intermediate results of our B-CAM, from left to right are the background localization prior $\bm{A}^{b}$, the object localization prior $\bm{A}^{o}$, the background localization score $\bm{S}^{B}$, the object localization score $\bm{S}^{O}$, the edge map of the predicted mask $\bm{Y}^{*}$ and ground truth $\bm{Y}$.}
\label{fig:vis}
\end{figure*}

Finally, to provide a visual interpretability of our B-CAM, some intermediate features are visualized in Fig.~\ref{fig:vis}, including the object/background localization priors $\bm{A}^{o}$, $\bm{A}^{b}$ and the localization scores $\bm{S}^{O}$, $\bm{S}^{B}$. The localization priors are visualized by the mean strength of $\bm{A}^o$ and $\bm{A}^b$. Specifically, the localization priors efficiently capture some representative background/object locations, which are then used to fuse the two aggregation features to represent pure-background and object samples. Then, the object score estimator which is trained based on these two aggregation features can generate better localization maps with less background activation. Moreover, our background score estimator can also generate precise background localization, which helps decide the final binary masks and bounding boxes. Though the boundary adherence is not very good compared with the ground truth mask due to the weakly supervised manner, our localization map captures most of the object parts in images.

\subsection{Limitations}
Though our B-CAM outperforms one-stage methods in most situations, some limitations still exist for our B-CAM. Firstly, as shown in Fig.~\ref{fig:vis}, though the mask of B-CAM can cover the object and avoid supervised threshold searching, some isolated regions may exist in the binary mask that cause performance decline. Secondly, the effect of our B-CAM is also influenced by the purity of image-level labels. Compared with datasets that annotate \textit{all objects} in an image (CUB-200/VOC-2012), the effect of our B-CAM is compromised for the datasets that only annotate \textit{the most conspicuous object} in an image (OpenImage). This is because the images with unlabeled object classes violate our Property 2 that images with certain objects must be the background of other classes, which introduces noise for background localization and results in the performance drop. This problem also causes our low performance in the large-scale ILSVRC dataset~\cite{IMAGENET}, which contains large amount of noisy image-level labels. These large amount noisy labels prevent our B-CAM from learning accuracy class-specific background classifier that is crucial to determine the final localization masks. We hope that future works can break through these limitations.

\section{Conclusion}
\label{sec:conclution}
In this paper, we propose B-CAM to extend the performance of WSOL methods by supplementing background awareness, which not only suppresses the excessive activation on background location but eliminates the need for threshold searching step. Experiments show that our proposed method achieves the best localization performance among one-stage methods. Our future work will extend our B-CAM into the downstream localization tasks, such as the weakly supervised semantic segmentation (WSSS) and weakly supervised object detection (WSOD). Moreover, we will also explore the applications of applying B-CAM in some specific fields such as lesion localization of medical images. 

\bibliographystyle{IEEEtran}
\bibliography{egbib}

\end{document}